\RequirePackage{fix-cm}
\documentclass[twocolumn]{svjour3}          
\smartqed  
\usepackage{xcolor}
\usepackage{tabularx}
\usepackage{multirow}
\usepackage{graphicx}
\usepackage{makecell}
\usepackage{amsmath}
\usepackage{color}
\usepackage{float}
\usepackage{rotating}
\usepackage{placeins}
\usepackage{amsfonts}
\usepackage{bm}

\usepackage[round,authoryear,comma,sort&compress,numbers]{natbib} 
\usepackage[colorlinks=true,linkcolor=blue,citecolor=black,urlcolor=blue]{hyperref}

\usepackage{diagbox}
\usepackage{cancel} 
\usepackage{mathrsfs} 
\usepackage{eqnarray} 

\usepackage{ctable} 

\usepackage{algpseudocode}%
\usepackage{listings}%
\usepackage{bbding}
\usepackage{bbm}
\usepackage[ruled]{algorithm2e}
\usepackage[caption=false, font=footnotesize]{subfig}
\usepackage{fontawesome}
\usepackage{array}
\newcommand{\PreserveBackslash}[1]{\let\temp=\\#1\let\\=\temp}
\newcolumntype{C}[1]{>{\PreserveBackslash\centering}p{#1}}
\newcolumntype{R}[1]{>{\PreserveBackslash\raggedleft}p{#1}}
\newcolumntype{L}[1]{>{\PreserveBackslash\raggedright}p{#1}}

\newcommand{\model}{SETR}
\newcommand{\modelFull}{\textit{SEgmentation TRansformer}}

\newcommand{\naiveModel}{SETR-\textit{Na\"ive}}
\newcommand{\pupModel}{SETR-\textit{PUP}}
\newcommand{\mlaModel}{SETR-\textit{MLA}}

\newcommand{\smallNaiveModel}{SETR-\textit{Na\"ive-Base}}
\newcommand{\smallMlaModel}{SETR-\textit{MLA-Base}}
\newcommand{\smallPupModel}{SETR-\textit{PUP-Base}}

\newcommand{\NaiveDeit}{SETR-\textit{Na\"ive-DeiT}}
\newcommand{\MlaDeit}{SETR-\textit{MLA-DeiT}}
\newcommand{\PupDeit}{SETR-\textit{PUP-DeiT}}
\newcommand{\MlaBeit}{SETR-\textit{MLA-BEiT}}
\newcommand{\PupBeit}{SETR-\textit{PUP-BEiT}}

\def\eg{\textit{e.g.}}
\def\ie{\textit{i.e.}}

\def\wrt{\textit{w.r.t}}

\definecolor{battleshipgrey}{rgb}{0.52, 0.52, 0.51}
\definecolor{capri}{rgb}{0.0, 0.75, 1.0}
\definecolor{mediumspringgreen}{rgb}{0.0, 0.98, 0.6}

\journalname{International Journal of Computer Vision}
\begin{document}

\title{Vision Transformers: From Semantic Segmentation to Dense Prediction
}

\titlerunning{Vision Transformers: From Semantic Segmentation to Dense Prediction}        

\author{
	Li Zhang$^{1\ast}$, \and
	Jiachen Lu$^{1\ast}$, \and 
        Sixiao Zheng$^{1\ast}$, \and 
        Xinxuan Zhao$^1$, \and 
	Xiatian Zhu$^2$, \and
        Yanwei Fu$^1$, \and 
        Tao Xiang$^2$, \and
        Jianfeng Feng$^1$, \and
        Philip H.S. Torr$^3$
}

\authorrunning{Li Zhang, et al.} 



\institute{
	Corresponding author: Li Zhang  \at
             \email{lizhangfd@fudan.edu.cn}          \\
$^{\ast}$ These authors contributed equally to this work. \\
$^1$ School of Data Science, Fudan University, Shanghai, China \\
$^2$ University of Surrey, Guildford, UK\\
$^3$ University of Oxford, Oxfordshire, UK
}

\date{July 2024}

\maketitle

\section{Introduction}
Since the introduction of Vision Transformers (ViTs) to image classification \cite{dosovitskiy2020image}, the landscape of visual representation learning has gradually shifted away from CNNs \cite{alexnet,vgg,resnet,googlenet,densenet}.
Inspired by this phenomenal success, {\em for the first time} we expand the application of ViTs
from image classification \cite{dosovitskiy2020image}
to more challenging dense prediction (\eg, semantic segmentation).
Specifically, we introduce {\bf\em SEgmentation TRansformer} (SETR) \cite{zheng2021rethinking} to replace the seminal fully convolutional network (FCN) \cite{fcn} with a ViT for visual representation learning.
It is shown that ViT can learn stronger {long-range dependency information} critical for semantic segmentation in unconstrained scene images \cite{segnet,denseaspp}, achieving superior semantic segmentation performances. 
Conceptually, this presents a {\em sequence-to-sequence} learning perspective, capable of learning discriminative visual representation at full receptive field per layer across all the image patches.
Subject to the computational budget, the resolution of convolutional feature maps of a FCN instead needs to reduce progressively, in order to learn more abstract and semantic visual concepts by gradually increased receptive fields. 
Indeed, there are a number of remedies introduced,
\eg, manipulating the convolution operation (large kernel sizes \cite{largekernel}, 
atrous convolutions \cite{holschneider1990real,deeplabv2},
and image/feature pyramids \cite{pspnet}),
integrating attention modules \cite{wang2018nonlocal,huang2018ccnet,li2019global},
and decomposed attention \cite{wang2020axial}.
Nonetheless, none of these can fully eliminate the limitation of FCN architecture in receptive field.
Instead, our reformulation offers an alternative to the dominating FCN design.
Crucially, it has been playing a timing, critical role in forming the recent surge of research on contextual visual representation learning \cite{liu2021swin,wang2021pvtv2,chen2021regionvit,li2021localvit,d2021convit}.

Specifically, SETR treats an input image as a sequence of {\em image patches} represented by learned patch embedding, and transforms the sequence with global self-attention modeling for discriminative feature representation learning. Concretely, we first decompose an image into a grid
of fixed-sized patches, forming a sequence of patches.
With a linear embedding layer applied to the 
flattened pixel vectors of every patch, we then obtain a 
sequence of feature embedding vectors
as the input to a ViT. Given the learned features from the encoder ViT, 
a decoder is then used 
to recover the original image resolution. 
Crucially there is {\em no} downsampling in spatial resolution but global context modeling at every layer of the encoder Transformer, thus offering a completely new perspective to the semantic segmentation problem.   
{Despite its strengths, the basic ViT architecture is less effective for broader dense prediction applications like object detection and instance segmentation. This limitation stems from its absence of a pyramidal structure, its lack of adequate local context processing and its considerable computational demands}

{To address this aforementioned limitation, in this work 
a novel {\bf\em Hierarchical Local-Global} (HLG) Transformers architecture is further introduced. 
Same as recent ViTs \cite{liu2021swin,chen2021regionvit,li2021localvit,chu2021twins,d2021convit},
we also borrow the pyramidal structure from CNNs \cite{resnet,densenet}.
This adaptation is essential for meeting the requirements of object detection and instance segmentation tasks, which benefit from hierarchical feature processing.
Critically, we design a generic HLG Transformer layer characterized 
by {\em local attention} within windows and {\em global-attention} across windows.
This configuration effectively compensates for the lack of local context processing in basic ViT models, which is crucial for accurately detecting smaller objects, while still leveraging the global context benefits of the vanilla ViT. The integration of these attention mechanisms enables the creation of a family of HLG Transformers that are not only superior in performance but also cost-effective, especially in lightweight configurations.}
Beyond semantic segmentation, we therefore further apply the proposed HLG Transformers to 
image classification and object detection problems, demonstrating the potentials of 
serving as a versatile ViT backbone.

Our {\bf contributions} with SETR are summarized as follows:
{\bf (1)} As an exemplar dense prediction problem, 
{we explore the potentials of self-attention mechanism for visual representation learning,}
offering an alternative to the dominating FCN design.
{\bf (2)} We exploit the ViT framework
to implement our image representation learning
by image sequentialization.
{\bf (3)} To extensively examine our representations,
we further introduce three decoder designs
at varying complexities.
Extensive experiments show that our SETR can 
learn superior visual representations as compared to 
different FCNs with and without attention modules, yielding 
new state of the art on ADE20K (50.28\%), Pascal Context (55.83\%) and competitive results on Cityscapes.
Particularly, our entry was ranked the {\em first} place in the highly competitive ADE20K test server leaderboard at the submission day.

Our preliminary work SETR~\cite{zheng2021rethinking} has been published in CVPR 2021. It is {\em the very first attempt} of exploring the Transformer for dense prediction tasks (\eg, semantic segmentation) in computer vision.
Encouragingly, it has been highly recognized  in the community and of great impact to the development of many follow-up works. 
In this paper, we further extend 
our preliminary version as below:
{\bf (1)} Extending the applications of 
ViTs from semantic segmentation to general dense prediction tasks (\eg, object detection, instance segmentation and semantic segmentation);
{\bf (2)}
Enhancing the conventional ViT architecture with building block of the new HLG Transformer layer
to allow more cost-effective and pyramidal visual representation learning;
{\bf (3)}
More extensive comparisons with recent stronger ViT variants on image classification,
object detection, and semantic segmentation benchmarks,
and demonstrating the superiority of our HLG Transformers over concurrent state-of-the-art alternatives.

\section{Related work}

\begin{figure*}[t]
\begin{minipage}[b]{.45\linewidth}
    \centering
    \subfloat[][]{\label{Genelecs:Genelec 8010 AP}\includegraphics[height=7cm, width=0.9\linewidth]{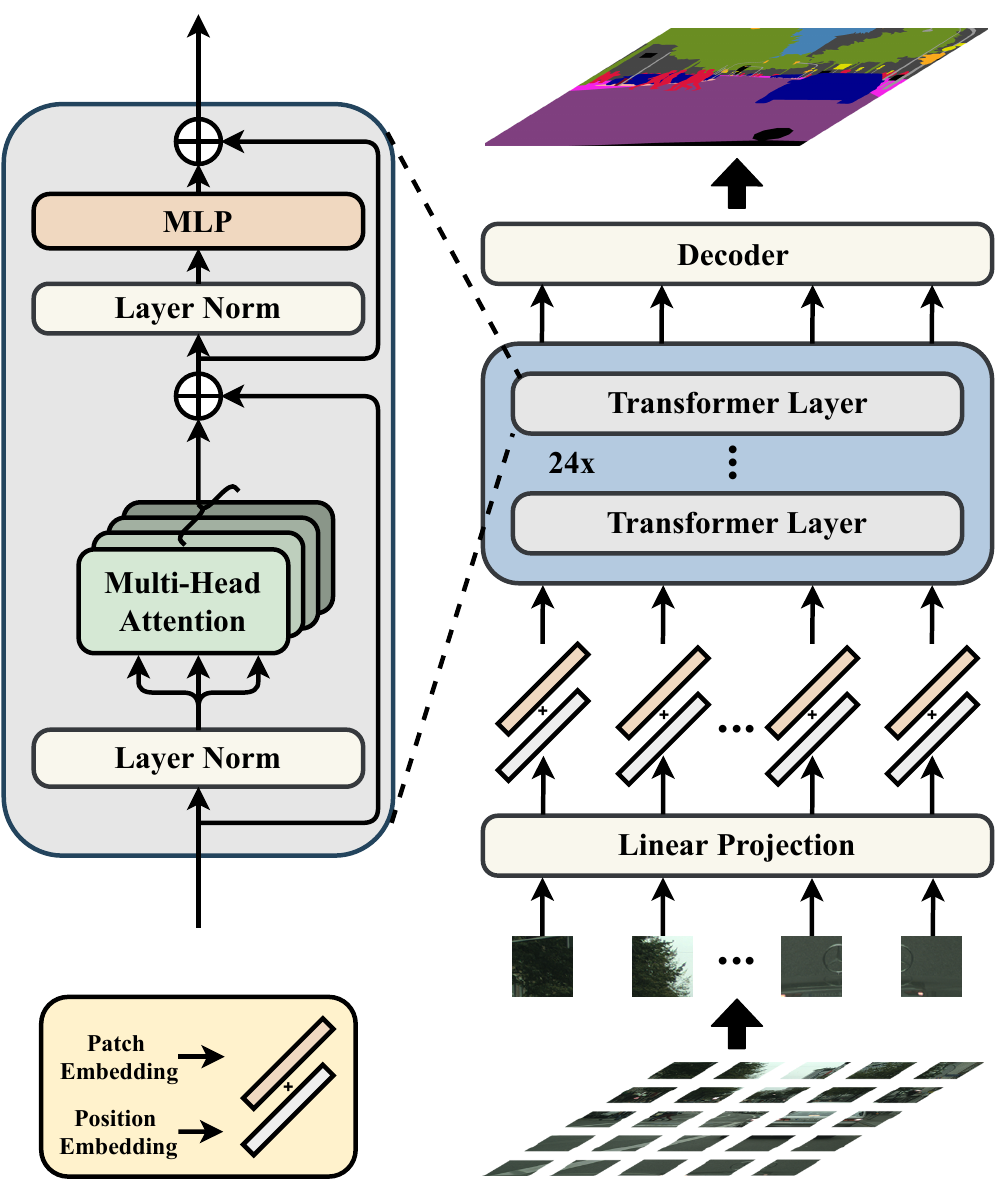}}
\end{minipage}
\medskip
\begin{minipage}[b]{.55\linewidth}
    \centering
    \subfloat[][]{\label{Genelecs:Genelec 8020 AP}\includegraphics[width=0.95\linewidth]{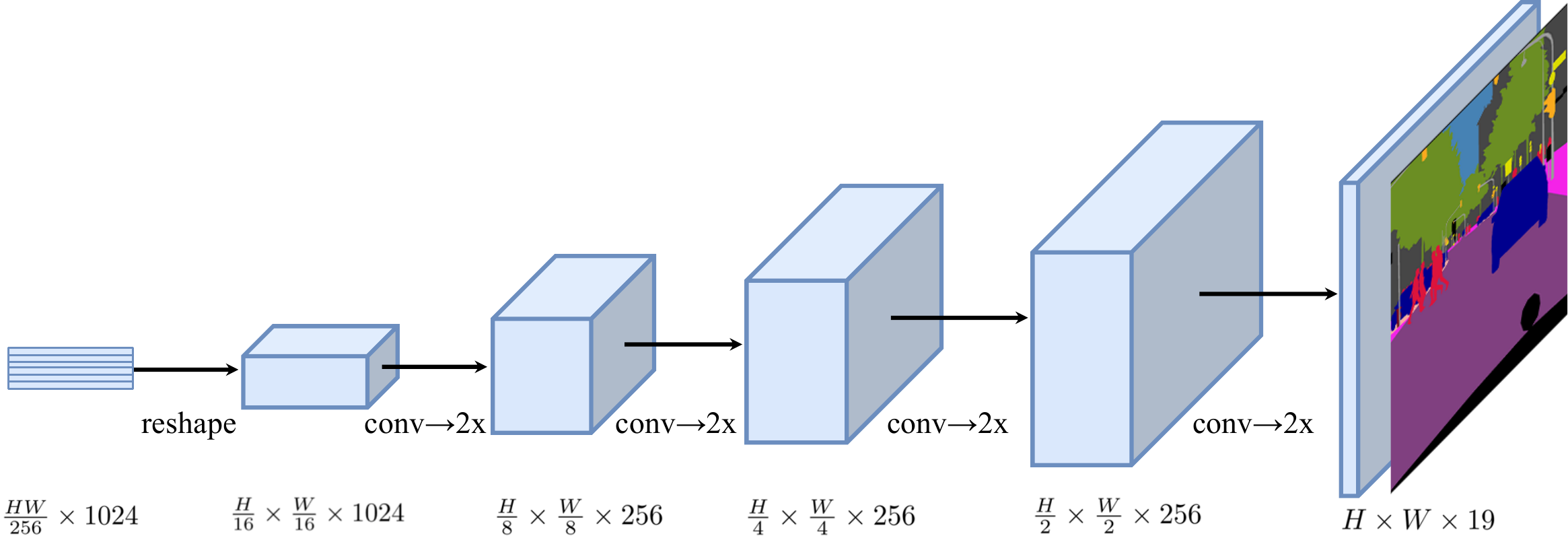}}
    
    \subfloat[][]{\label{Genelecs:Genelec 8030 AP}\includegraphics[width=0.95\linewidth]{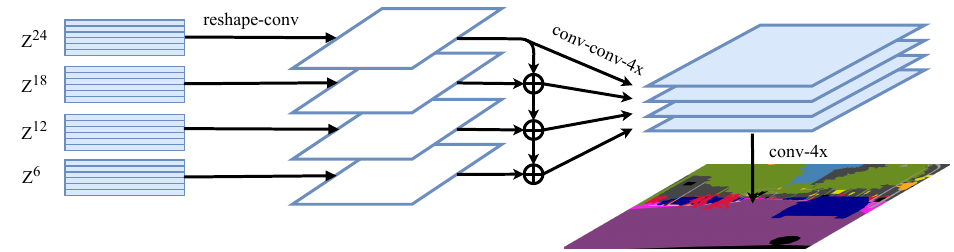}}
\end{minipage} 

\caption{\textbf{Schematic illustration of the proposed 
\modelFull~(\model)} (a).
We first split an image into fixed-size patches, linearly embed each of them, add position embeddings, and feed the resulting sequence of vectors to a standard Transformer encoder. 
To perform pixel-wise segmentation, we introduce different decoder designs:
(b) progressive upsampling (resulting in a variant called \pupModel); and
(c) multi-level feature aggregation (a variant called \mlaModel).
}
\label{fig:SETR_transformer}
\end{figure*}

\subsection{Semantic Segmentation}
Semantic segmentation has been significantly boosted by the development of deep neural networks. By removing  fully connected layers, a fully convolutional network (FCN)~\cite{fcn} can achieve pixel-wise predictions. While the predictions of FCN are relatively coarse, several CRF/MRF~\cite{chen2015semantic,liu2015semantic,zheng2015conditional} based approaches are developed to help refine the coarse predictions. To address the inherent tension between semantics and location~\cite{fcn}, coarse and fine layers need to be aggregated for both the encoder and decoder. This leads to different variants of the encoder-decoder structures~\cite{unet,deconvnet,segnet} for multi-level feature fusion.

Many recent  efforts have been focused on addressing the limited receptive field/context modeling problem in FCN.  To enlarge the receptive field, DeepLab~\cite{deeplabv1} and Dilation~\cite{dilation} introduce the dilated convolution. Alternatively, context modeling is the focus of PSPNet~\cite{pspnet} and DeepLabV2~\cite{deeplabv2}. The former proposes the PPM module to obtain different region's contextual information while the latter develops ASPP module that adopts pyramid dilated convolutions with different dilation rates. Decomposed large kernels~\cite{largekernel} are also utilized for context capturing.
Recently, attention based models are popular for capturing long range context information. PSANet~\cite{psanet} develops the pointwise spatial attention module for dynamically capturing the long range context. DANet~\cite{DAnet} embeds both spatial attention and channel attention. CCNet~\cite{huang2018ccnet} alternatively focuses on economizing the heavy computation budget introduced by full spatial attention. DGMN~\cite{zhang2020dynamic} builds a dynamic graph message passing network for scene modeling and it can significantly reduce the computational complexity. Note that all these approaches are still based on FCNs where the feature encoding and extraction part are based on classical ConvNets like VGG~\cite{vgg} and ResNet~\cite{resnet}. In this work, we alternatively rethink the semantic segmentation task from a different perspective.

\subsection{Vision Transformers}
Transformers have revolutionized  machine translation and NLP~\cite{vaswani2017attention,devlin2018bert,dai2019transformer,Yang2019xlnet}. Recently, there are also some explorations for the usage of Transformer structures in image recognition. Non-local network~\cite{wang2018nonlocal} appends Transformer style attention onto the convolutional backbone. AANet~\cite{bello2019attention} mixes convolution and self-attention for backbone training. LRNet~\cite{hu2019local} and stand-alone networks~\cite{ramachandran2019stand} explore local self-attention to avoid the heavy computation brought by global self-attention. SAN~\cite{zhao2020san} explores two types of self-attention modules. Axial-Attention~\cite{wang2020axial} decomposes the global spatial attention into two separate axial attentions such that the computation is largely reduced. Apart from these pure Transformer based models, there are also CNN-Transformer hybrid ones. DETR~\cite{carion2020end} and the following deformable version utilize Transformer for object detection where Transformer is appended inside the detection head. STTR~\cite{li2020revisiting} and LSTR~\cite{liu2020lane} adopt Transformer for disparity estimation and lane shape prediction respectively. Recently, ViT~\cite{dosovitskiy2020image} is the first work to show that a pure Transformer based image classification model can achieve the state-of-the-art. It provides direct inspiration to exploit Transformer based encoder design for semantic segmentation.

The most related work is \cite{wang2020axial} which  also leverages attention for image segmentation. However, there are several key differences. First, though convolution is completely removed in \cite{wang2020axial} as in our SETR, their model still follows the conventional FCN design in that spatial resolution of feature maps is reduced progressively. In contrast, our prediction model keeps the same spatial resolution throughout and thus represents a step-change in model design. 
Second, to maximize the scalability on modern hardware accelerators and facilitate easy-to-use, we stick to the standard self-attention design. Instead, \cite{wang2020axial} adopts a specially designed axial-attention  \cite{ho2019axial} which is less scalable to standard computing facilities. Our model is also superior in segmentation accuracy (see Section~\ref{sec:exp}). 

{Despite the success of ViTs~\cite{dosovitskiy2020image} in coarse classification tasks, there is a need for more cost-effective variants and the integration of local and global context relationships for more challenging dense prediction tasks.
For example, Transformer in transformer~\cite{han2021transformer} first introduce Transformer block to local patch, but it's still kept in single-stage manner.
PVT~\cite{wang2021pyramid} first introduces spatial reduction attention in a pyramid architecture. 
CvT~\cite{wu2021cvt} similarly coalesces ViTs with convolutional inductive bias.
CrossViT~\cite{chen2021crossvit} instead leverages interaction between large-patch and small-patch branches.
Swin~\cite{liu2021swin} shifts the local attention windows to
capture global cross-window connection across layers.
Rather than employing pure local attention, DiNAT~\cite{hassani2023neighborhood} applies dilation to local attention and utilizes different dilation rates to achieve global context learning.
Recently, \citep{li2107local, chu2021twins, song2022all} have directly integrated local and global attention in their attention mechanisms. Local-Global~\cite{li2107local} incorporates local attention windows at three different levels. Twins~\cite{chu2021twins} combines locally-grouped attention with globally sub-sampled attention across the group. GLAM~\cite{song2022all} employs channel-wise attention, although it still retains the costly globally self-attention.
}

{
While the local attention in these methods remains confined to small windows, which can limit the capture of long-range context, our Hierarchical-Local-Global Transformers utilize dilation in the local windows. This hierarchical approach enables the model to capture both local and long-range information across multiple layers.
}






\section{Method}
\begin{figure*}[h]
  \centering
  \includegraphics[width=1.0\linewidth]{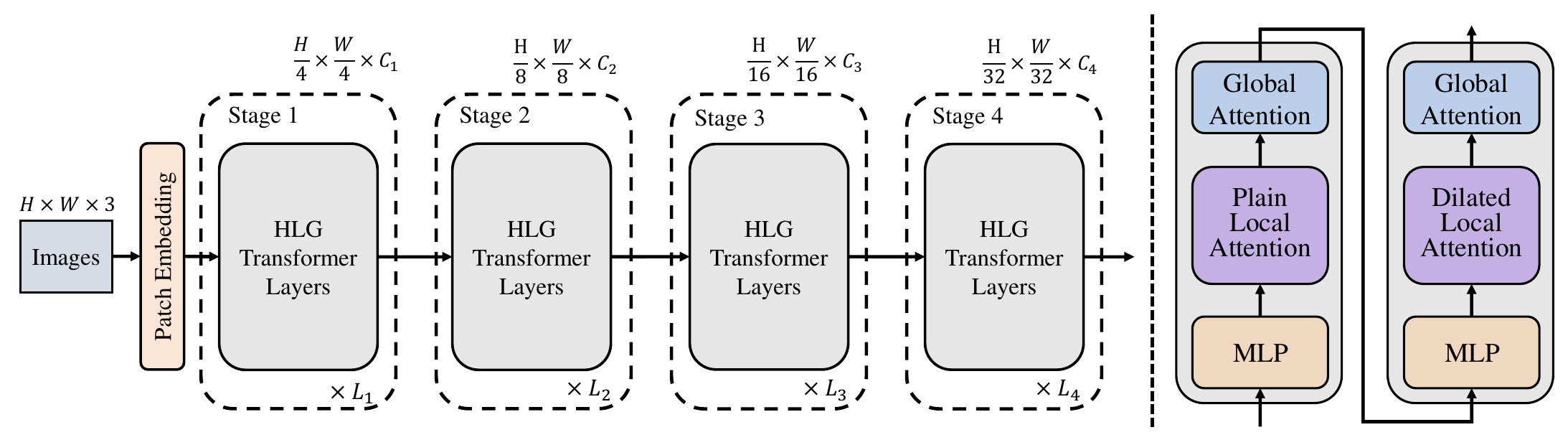}
  \caption{
  {\bf Left:} Hierarchical local-global Transformers backbone architecture. 
  {\bf Right:} Successive hierarchical local-global Transformer layer.}
  \label{fig:hlg_architecture}
\end{figure*}
\subsection{FCN-based Semantic Segmentation}
In order to contrast with our new model design, let us first revisit the conventional FCN \cite{fcn} for image semantic segmentation. An FCN encoder consists of a stack of sequentially connected convolutional layers. The first layer takes as input the image, denoted as $H \times W \times 3$ with 
 $H \times W $ specifying the image size in pixels.  
The input of subsequent layer $i$ is a three-dimensional tensor sized $h \times w \times d$, where $h$ and $w$ are spatial dimensions of feature maps, and $d$ is the feature/channel dimension.
Locations of the tensor in a higher layer are computed based on the locations of tensors of all lower layers they are connected to via layer-by-layer convolutions, which are defined as their {\em receptive fields}.
Due to the locality nature of convolution operation, the receptive field increases linearly along the depth of layers, conditional on the kernel sizes (typically $3 \times 3$).
As a result, only higher layers with big receptive fields can model long-range dependencies
in this FCN architecture.
However, it is shown that the benefits of adding more layers
would diminish rapidly once reaching certain depths \cite{resnet}. Having limited receptive fields for context modeling is thus an intrinsic limitation of the vanilla FCN architecture. 

Recently, a number of state-of-the-art methods \cite{zhang2019dual,huang2018ccnet,zhang2020dynamic}
suggest that combing FCN with attention mechanism is a more effective strategy for learning long-range 
contextual information.
These methods limit the attention learning
to higher layers with smaller input sizes alone due to its quadratic complexity
\wrt the pixel number of feature tensors.
This means that dependency learning on lower-level feature tensors is lacking, leading to sub-optimal representation learning.
To overcome this limitation, we propose a pure self-attention based encoder, named {\em SEgmentation TRansformers} (SETR). 

\begin{figure*}[!htb]
  \centering
  \includegraphics[width=1.0\linewidth]{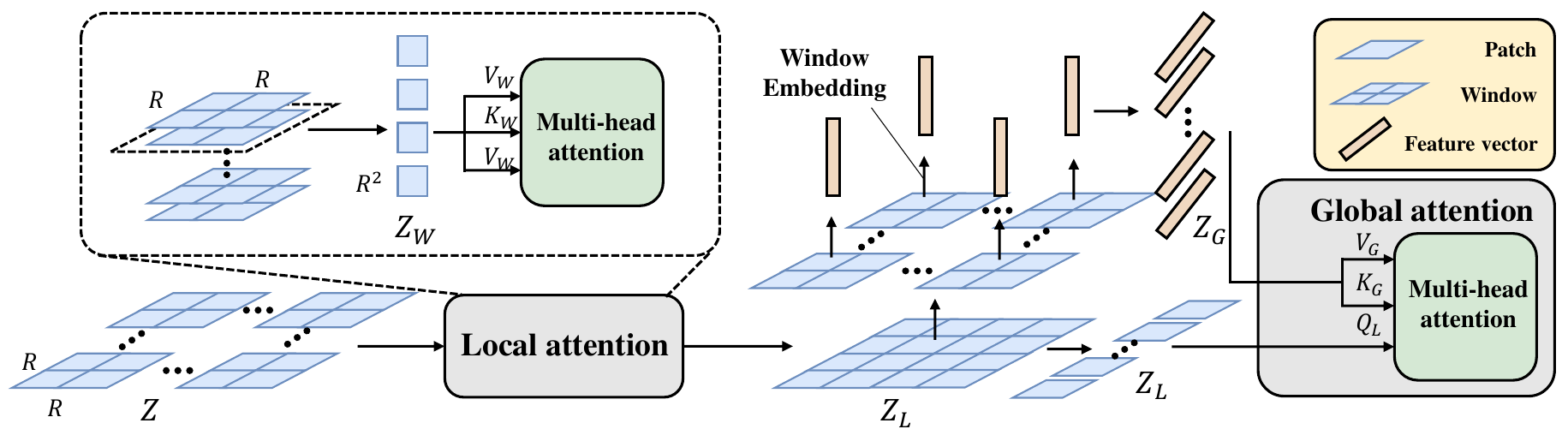}
  \caption{\textbf{Local-Global attention mechanism.} Local attention (in dash line) is applied within each window. 
  Global attention is applied between global feature $Z_G$ and local feature $Z_L$.
  }
  \label{fig:hlg_block}
\end{figure*}
\subsection{Segmentation Transformers (SETR)}
\noindent \textbf{Single-stage Transformer} 
Given the 1D embedding sequence $E$ as input,
a pure transformer based encoder is employed to learn feature representations. 
Concretely, the Transformer, as depicted in  Figure~\ref{fig:SETR_transformer}(a), accepts a 1D sequence of feature embeddings $Z \in \mathbb{R}^{L\times C}$ as input, $L$ is the length of sequence, $C$ is the hidden channel size. 
This means each transformer layer has a global 
 receptive field, solving the limited receptive field problem of existing FCN encoder once and for all.
The transformer encoder consists of $L_e$ layers of multi-head
self-attention (MSA) and Multilayer Perceptron (MLP) blocks \cite{velivckovic2017graph}
(Figure \ref{fig:SETR_transformer}(a)).
At each layer $l$, the input to self-attention is in a triplet of
(\texttt{query}, \texttt{key}, \texttt{value})
computed from the input $Z^{l-1} \in \mathbb{R}^{L\times C}$ as:
\begin{align}\footnotesize
    \text{query} = {Z^{l-1}} \textbf{W}_Q, \; 
    \text{key} = {Z^{l-1}} \textbf{W}_K, \;
    \text{value} = {Z^{l-1}} \textbf{W}_V,
\end{align}
where $\textbf{W}_Q$/$\textbf{W}_K$/$\textbf{W}_V  \in \mathbb{R}^{C\times d}$
are the learnable parameters of three linear projection layers
and $d$ is the dimension of (\texttt{query}, \texttt{key}, \texttt{value}).
Self-attention (SA) is then formulated as:
\begin{equation}\tiny
    SA(Z^{l-1}) = Z^{l-1} + 
    \operatorname{softmax}(\frac{Z^{l-1} \textbf{W}_Q (Z^{l-1} \textbf{W}_K)^\top}{\sqrt{d}}) (Z^{l-1} \textbf{W}_V).
    \label{eq:attn}
\end{equation}
MSA is an extension with $m$ independent SA operations
and project their concatenated outputs:
$MSA(Z^{l-1}) = [SA_1(Z^{l-1});~SA_2(Z^{l-1}); ~\cdots;~SA_m(Z^{l-1})]\textbf{W}_O$, where $\textbf{W}_O \in \mathbb{R}^{md \times C} $. $d$ is typically set to $C/m$.
The output of MSA is then transformed 
by an MLP block with residual skip as the layer output as:
\begin{equation}\footnotesize
    Z^l = MSA(Z^{l-1}) + MLP(MSA(Z^{l-1})) \in \mathbb{R}^{L \times C}.
\end{equation}
Note, layer norm is applied before MSA and MLP blocks
which is omitted for simplicity.
We denote $\{Z^1,~Z^2,~\cdots,~Z^{L_e}\}$ as the features of transformer layers.

\noindent \textbf{Image to sequence} 
To suit the 1D embedding sequnce as input, there thus exists a mismatch between 2D image and 1D sequence. 
Image sequentialization is thus needed to convert an input image $x\in \mathbb{R}^{H\times W\times 3}$ into $Z$.

A straightforward way for image sequentialization is 
to flatten the image pixel values into a 1D vector with size of $3HW$.
For a typical image sized at $480 (H)\times480(W)\times3$,
the resulting vector will have a length of 691,200. 
Given the quadratic model complexity of Transformer,
it is not possible that such high-dimensional vectors 
can be handled in
both space and time. 
Therefore tokenizing every single pixel as input to our transformer is out of the question.  

In view of the fact that a typical encoder designed for semantic segmentation would downsample a 2D image $x\in \mathbb{R}^{H\times W\times 3}$ into a feature map $x_f\in \mathbb{R}^{\frac{H}{16}\times \frac{W}{16}\times C}$, we  thus decide to set the transformer input sequence length $L$ as $\frac{H}{16} \times \frac{W}{16} = \frac{HW}{256}$. 
This way, the output sequence of the transformer can be simply reshaped to the target feature map $x_f$.

To obtain the $\frac{HW}{256}$-long input sequence, we divide an image $x\in \mathbb{R}^{H\times W\times 3}$ into a grid of $\frac{H}{16}\times\frac{W}{16}$ patches uniformly,
and then flatten this grid into a sequence.
By further mapping each vectorized patch $p$ into a latent $C$-dimensional embedding space
using a linear projection function $f$:
$p \longrightarrow e \in \mathbb{R}^C$, 
we obtain a 1D sequence of patch embeddings
for an image $x$.
To encode the patch spacial information,
we learn a specific embedding $p_i$ for every location $i$
which is added to $e_i$ to form the final sequence input $E = \{e_1+p_1,~e_2+p_2,~\cdots,~e_L+p_L\}$. This way, spatial information is kept despite the orderless self-attention nature of transformers. 

\begin{figure}[!ht]\centering
\includegraphics[width=1.0\linewidth]{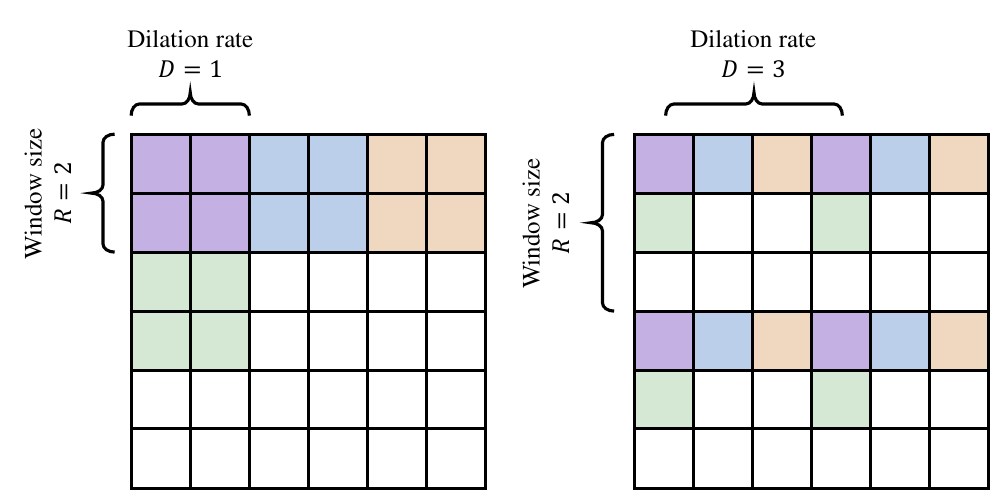}
\caption{
{\bf Left:} plain local attention;
{\bf Right:} dilated local attention.}
\label{fig:hlg_dis}
\end{figure}
\subsection{Hierarchical Local-Global Transformer}

As a conventional ViT~\cite{dosovitskiy2020image} learns feature representation of a single resolution with quadratic computation complexity \wrt~the input image size, 
it is not appropriate to be a backbone network to support various fine downstream tasks, \eg, object detection and instance segmentation.
In this paper, 
we introduce a family of {\bf Hierarchical Local-Global} (HLG) Transformers in a pyramidal structure, serving as a general-purpose backbone network.
An overview of our 
HLG architecture is depicted in Figure \ref{fig:hlg_architecture}.
A HLG model consists of four stages for multi-scale feature representation learning. 
Specifically, an input image is first fed to the patch embedding module and downsampled by overlapping patch divisions, followed by $L_{i}$ HLG layers with $i$ the stage index. All stages share the same structure including one patch embedding module and multiple encoding modules. The output size of each stage $x_f^0,x_f^1,x_f^2,x_f^3$ is designated to be $ \frac{H}{4} \times \frac{W}{4}$, $\frac{H}{8} \times \frac{W}{8}$, $ \frac{H}{16} \times \frac{W}{16} $, $ \frac{H}{32} \times \frac{W}{32} $ sequentially.
Next, we describe the key design of the proposed HLG Transformer layer.

\noindent \textbf{Local-Global attention}
An illustration of our local-global attention is given at Figure \ref{fig:hlg_block}.

\noindent \textbf{Local attention}
Given an input feature map $Z$ of size $N_h\times N_w $, it is first divided into $ \frac{N_h\times N_w}{R^2} $ windows in a non-overlapping manner, where $ R \times R $ represents the window size.
The multi-head self-attention is then applied within each local window $Z_W\in\mathbb{R}^{R\times R\times C}$.
The attention operation can be expressed as:
\begin{equation}
\label{equ:local_qkv}
    Q_W=Z_W\mathbf{W}^{(L)}_Q, K_W=Z_W\mathbf{W}^{(L)}_K,
    V_W=Z_W\mathbf{W}^{(L)}_V
\end{equation}
\begin{equation}
\label{equ:local_attn}
    \text{SA}(Z_W)=Z_W + \text{softmax}\left(
    \frac{Q_W K^{\top}_W}{\sqrt{d}}+B_W
    \right)V_W,
\end{equation}
where $\mathbf{W}^{(L)}_Q,\mathbf{W}^{(L)}_K,\mathbf{W}^{(L)}_V\in\mathbb{R}^{C\times d}$ are learnable parameters of three linear projection layers shared by all windows and $d$ is the dimension of $Q_W, Q_K, Q_V$. 
$B_W\in \mathbb{R}^{R^2\times R^2}$ is local relative positional encoding.
Then we assemble each window feature back to obtain the local feature map $ Z_L \in\mathbb{R}^{N_h\times N_W\times C}$.
This {\em plain local attention} can be achieved by non-overlapping window division with focus on information exchange between spatially adjacent patches (Figure \ref{fig:hlg_dis} left).
This scheme is however limited to short range interaction within each individual window.
To enable long range dependency discovery without extra computation cost, we further introduce a {\em dilated local attention} mechanism for stronger representation learning (Figure \ref{fig:hlg_dis} right). 
In particular, we form anther type of window by sampling patches
in a spaced manner, \eg, sampling a patch every three positions as illustrated
in (Figure \ref{fig:hlg_dis}).
This allows to learn a relatively longer range dependency using efficient local attention.

\noindent \textbf{Global attention}
Following local feature map $Z_L$, we further design an efficient global attention for holistic context learning.
To that end, we introduce a {\em Window Embedding} operation that extracts an 1-dimensional feature vector $\mathbb{R}^{1\times 1\times C}$ per window.
Window embedding operations can be implemented by depth-wise convolution~\cite{chollet2017xception}, average pooling or max pooling with kernel size and stride equal to the window size.
Empirically we find average pooling is generally effective for downstream tasks.
More details will be discussed in the experiments.
As a result, a {\em compact} global feature map $Z_G\in\mathbb{R}^{\frac{N_h}{R}\times \frac{N_w}{R}}\times C$ can be obtained.
Global attention is then applied to feature map $Z_L$ and $Z_G$.
This enables efficient global attention by querying 
$Z_L$ towards $Z_G$ formulated as:
\begin{equation}\footnotesize
\label{equ:global_qkv}
    Q_L=Z_L\mathbf{W}^{(G)}_Q,\quad K_G=Z_G\mathbf{W}^{(G)}_K,\quad
    V_G=Z_G\mathbf{W}^{(G)}_V
\end{equation}
\begin{equation}\footnotesize
\label{equ:global_attn}
    \text{SA}(Z_L, Z_G)=Z_L + \text{softmax}\left(
    \frac{Q_L K_G^\top}{\sqrt{d}}+B_G
    \right)V_G,
\end{equation}
where $\mathbf{W}^{(G)}_Q,\mathbf{W}^{(G)}_K,\mathbf{W}^{(G)}_V\in\mathbb{R}^{C\times d}$ are learnable parameters of three linear projection layers and $d$ is the dimension of $Q_W, Q_K, Q_V$. $B_G\in \mathbb{R}^{(N_h\times N_w)\times (\frac{N_h}{R}\times \frac{N_w}{R})}$ is global relative positional encoding.

\noindent \textbf{Parameter and computation sharing}
For further efficiency, 
we propose to share the parameters and computation between local and global attention.
Concretely, all the parameters of local and global attention are shared as
\begin{equation}
\mathbf{W}^{(L)}_Q=\mathbf{W}^{(G)}_Q,\mathbf{W}^{(L)}_K=\mathbf{W}^{(G)}_K,\mathbf{W}^{(L)}_V=\mathbf{W}^{(G)}_V
\end{equation}
Computation sharing is non-trivial.
To that end, we replace $Z_L$ of Eq.~\eqref{equ:global_qkv} with the output of Eq.~\eqref{equ:local_attn} as:
\begin{equation}\footnotesize
\begin{split}
    \notag Q_L &= \left(Z_W + \text{softmax}\left(
    \frac{Q_W K^{\top}_W}{\sqrt{d}}+B_W
    \right)V_W \right)\mathbf{W}^{(G)}_Q\\
    \notag&=Z_WW_Q^{(L)} + \text{softmax}\left(
    \frac{Q_W K^{\top}_W}{\sqrt{d}}+B_W
    \right)V_W\mathbf{W}^{(G)}_Q
\end{split}
\end{equation}
As a result, we achieve the the first term by reusing Eq.~\eqref{equ:local_qkv}, and the second term by a computation-friendly depth-with convolution as
\begin{equation}
    Q_L = Q_W + \text{DWConv}\left(V_W\right)
\end{equation}
where $Q_W$ and $K_W$ are shared with local attention.

\noindent \textbf{DWMLP}
Following \cite{sandler2018mobilenetv2,yang2022moat}, we implement our MLP with a depth-wise convolution and squeeze-and-excitation~\cite{hu2018squeeze}.
We place the MLP before attention 
so that downsampling can be achieved simply by setting the stride of MLP (first layer in each stage) as 2.

\noindent \textbf{Hierarchical Local-Global Transformer layer}
To facilitate synergistic interaction without extra complexity and computational cost, we intertwine the plain and dilated local attention sequentially.
Both use the same window size.
This attention cascading design results in our Hierarchical Local-Global (HLG) Transformer layer (Figure~\ref{fig:hlg_architecture} right).

\noindent \textbf{Architecture design}
\begin{table*}[htb]
\caption{
Architecture specifications of HLG. $ Conv3 \times 3 $ means the convolution with a kernel size of $ 3 \times 3$, $ Conv1 \times 1 $ denotes the convolution with a kernel size of $ 1 \times 1 $, and $ DWConv3 \times 3 $ represents the depth-wise convolution with a kernel size of $ 3 \times 3 $.
}
\centering
\setlength{\tabcolsep}{1pt} 
\renewcommand{\arraystretch}{1.2} 
\begin{tabular*}{\textwidth}{@{\extracolsep\fill}cccccc}
\toprule
 Output & HLG-Mobile & HLG-Tiny & HLG-Small & HLG-Medium & HLG-Large \\
 \hline
 
 \multirow{5}{*}{ $ 56 \times 56 $ } 
 & \multicolumn{5}{c}{ $ Conv3\times3 (stride=2), Conv1\times1, Conv3\times3(stride=2) $ } \\ 
 \cmidrule{2-6}
 & $ {\begin{bmatrix} C_1=48 \\ H_1=2 \\ R_1=7 \\ D_1=8 \\ \end{bmatrix}} \times 2 $ 
 & $ {\begin{bmatrix} C_1=64 \\ H_1=2 \\ R_1=7 \\ D_1=8 \\ \end{bmatrix}} \times 2 $ 
 & $ {\begin{bmatrix} C_1=96 \\ H_1=3 \\ R_1=7 \\ D_1=8 \\ \end{bmatrix}} \times 2 $ 
 & $ {\begin{bmatrix} C_1=96 \\ H_1=3 \\ R_1=7 \\ D_1=8 \\ \end{bmatrix}} \times 2 $  
 & $ {\begin{bmatrix} C_1=128 \\ H_1=4 \\ R_1=7 \\ D_1=8 \\ \end{bmatrix}} \times 2 $  \\
 \midrule
 
 \multirow{5}{*}{ $ 28 \times 28 $ } & \multicolumn{5}{c}{ $ DWConv3\times3(stride=2) $ } \\ 
 \cmidrule{2-6}
 & $ {\begin{bmatrix} C_2=96 \\ H_2=4 \\ R_2=7 \\ D_2=4 \\ \end{bmatrix}} \times 2 $ 
 & $ {\begin{bmatrix} C_2=128 \\ H_2=4 \\ R_2=7 \\ D_2=4 \\ \end{bmatrix}} \times 2 $ 
 & $ {\begin{bmatrix} C_2=192 \\ H_2=6 \\ R_2=7 \\ D_2=4 \\ \end{bmatrix}} \times 2 $ 
 & $ {\begin{bmatrix} C_2=192 \\ H_2=6 \\ R_2=7 \\ D_2=4 \\ \end{bmatrix}} \times 2 $  
 & $ {\begin{bmatrix} C_2=256 \\ H_2=8 \\ R_2=7 \\ D_2=4 \\ \end{bmatrix}} \times 2 $  \\
 \midrule
 
 \multirow{5}{*}{ $ 14 \times 14 $ } & \multicolumn{5}{c}{$ DWConv3\times3(stride=2) $} \\ 
 \cmidrule{2-6}
 & $ {\begin{bmatrix} C_3=192 \\ H_3=8 \\ R_3=7 \\ D_3=2 \\ \end{bmatrix}} \times 2 $ 
 & $ {\begin{bmatrix} C_3=256 \\ H_3=8 \\ R_3=7 \\ D_3=2 \\ \end{bmatrix}} \times 6 $ 
 & $ {\begin{bmatrix} C_3=384 \\ H_3=12 \\ R_3=7 \\ D_3=2 \\ \end{bmatrix}} \times 6 $ 
 & $ {\begin{bmatrix} C_3=384 \\ H_3=12 \\ R_3=7 \\ D_3=2 \\ \end{bmatrix}} \times 14 $  
 & $ {\begin{bmatrix} C_3=512 \\ H_3=16 \\ R_3=7 \\ D_3=2 \\ \end{bmatrix}} \times 14 $  \\
 \midrule
 
 \multirow{5}{*}{ $ 7 \times 7 $ } & \multicolumn{5}{c}{$ DWConv3\times3(stride=2) $} \\ 
 \cmidrule{2-6}
 & $ {\begin{bmatrix} C_4=384 \\ H_4=16 \\ R_4=7 \\ D_4=1 \\ \end{bmatrix}} \times 2 $ 
 & $ {\begin{bmatrix} C_4=512 \\ H_4=16 \\ R_4=7 \\ D_4=1 \\ \end{bmatrix}} \times 2 $ 
 & $ {\begin{bmatrix} C_4=768 \\ H_4=24 \\ R_4=7 \\ D_4=1 \\ \end{bmatrix}} \times 2 $ 
 & $ {\begin{bmatrix} C_4=768 \\ H_4=24 \\ R_4=7 \\ D_4=1 \\ \end{bmatrix}} \times 2 $  
 & $ {\begin{bmatrix} C_4=1024 \\ H_4=32 \\ R_4=7 \\ D_4=1 \\ \end{bmatrix}} \times 2 $  \\
 \bottomrule
\end{tabular*}

\label{tab:hlg_architecture}
\end{table*}

Table \ref{tab:hlg_architecture} summarizes the family of our HLG Transformers with a variety of capacities, including HLG-Mobile, -Tiny, -Small, -Medium, -Large. The hyper-parameters are explained as follows:
\begin{itemize}
    \item $ C_i $: The embedding dimension of each stage $i$;
    \item $ H_i $: The number of head in stage $ i $;
    \item $ R_i $: The window size of local attention;
    \item $ D_i $: The dilation rate of local attention.
\end{itemize}

\subsection{Decoder Designs}
\label{sec:decoder}
\noindent \textbf{Singl-stage Transformer segmentation decoder}
To evaluate the effectiveness of SETR's encoder feature representations $Z$, we introduce three different decoder designs to perform pixel-level segmentation.
As the goal of the decoder is to generate the segmentation results in the original 2D image space $(H \times W)$, we need to reshape the encoder's features (that are used in the decoder),
$Z$, from a 2D shape of $\frac{HW}{256} \times C$
to a standard 3D feature map $\frac{H}{16} \times \frac{W}{16} \times C$. Next, we briefly describe the three decoders. 

\noindent \textbf{(1) Naive upsampling (Naive)}
This naive decoder 
first projects the transformer feature $Z^{L_e}$ to the dimension of category number (\eg, 19 for experiments on Cityscapes).
For this we adopt a simple 2-layer network with architecture:
$1 \times 1$ conv + sync batch norm (w/ ReLU) + $1 \times 1$ conv.
After that, we
simply bilinearly upsample the output to the full image resolution,
followed by a classification layer with pixel-wise cross-entropy loss.
When this decoder is used,
we denote our model as \naiveModel.

\noindent \textbf{(2) Progressive UPsampling (PUP)}
Instead of one-step upscaling which may introduce  noisy
predictions, we consider a {\em progressive upsampling} strategy 
that alternates conv layers and upsampling operations.
To maximally mitigate the adversarial effect,
we restrict upsampling to 2$\times$.
Hence, a total of 4 operations are needed for reaching 
the full resolution from $Z^{L_e}$ with size $\frac{H}{16} \times \frac{W}{16}$.
More details of this process are given in Figure \ref{fig:SETR_transformer}(b).
When using this decoder,
we denote our model as \pupModel.

\noindent \textbf{(3) Multi-Level feature Aggregation (MLA)}
The third design is characterized by multi-level feature aggregation (Figure \ref{fig:SETR_transformer}(c))
in similar spirit of feature pyramid network \cite{fpn,kirillov2019panoptic}.
However, our decoder is  fundamentally different
because the feature representations $Z^l$ of every SETR's layer share the same resolution without a pyramid shape.

Specifically, we take as input the feature representations
$\{ Z^m\}$ ($m \in \{{\frac{L_e}{M}}, {2\frac{L_e}{M}}, \cdots, {M\frac{L_e}{M}} \}$)
from $M$ layers uniformly distributed across the layers with step $\frac{L_e}{M}$ to the decoder.
$M$ streams are then deployed, with each focusing on one specific
selected layer.
In each stream, 
we first reshape the encoder's feature
$Z^{l}$ from a 2D shape of $\frac{HW}{256} \times C$
to a 3D feature map $\frac{H}{16} \times \frac{W}{16} \times C$.
A 3-layer (kernel size $1\times 1$, $3\times 3$, and $3 \times 3$) network is applied with the feature channels 
halved at the first and third layers respectively,
and the spatial resolution upscaled $4\times$ by bilinear operation
after the third layer.
To enhance the interactions across different streams, we introduce a top-down aggregation design via element-wise addition after the first layer.
An additional $3 \times 3$ conv is applied after the element-wise additioned feature.
After the third layer, we obtain the fused feature from all the streams via channel-wise concatenation which is then bilinearly upsampled $4\times$ to the full resolution.
When using this decoder, we denote our model as \mlaModel.

\noindent \textbf{Hierarchical Transformer segmentation decoder}
We detail the integration of HLG for segmentation.
To align with the input format (\ie, a spatial shape of $\frac{H}{16}\times\frac{W}{16}$) of SETR decoder,
we interpolate the features of all four stages
to this shape, followed by concatenating and transforming them
into a single tensor.
The decoder uses a pair of HLG Transformer layers with plain and dilated local attention
with the window size $R$ set to $8$.
The segmentation result is obtained by Progressive UPsampling (PUP).

\noindent \textbf{Hierarchical Transformer decoder}
In addition to adapting the hierarchical Transformer architecture to single-stage transformers, the pyramidal structure facilitates the use of the same decoders that are common in CNN-based methods. We employ UperNet \cite{upernet} as the pyramidal segmentation decoder, RetinaNet \cite{lin2017focal} for object detection, and Mask-RCNN \cite{he2017mask} as the decoder for instance segmentation.
\section{Experiments}
\label{sec:exp}

\begin{table}[!htb]
\caption{Configuration of SETR backbone variants.}
\label{tab:config}
\centering
    \begin{tabular*}{\columnwidth}{@{\extracolsep\fill}cccc}
        \toprule
        Model &  T-layers & Hidden size & Att head \\
        \midrule
        T-Base &12 & 768 & 12 \\
        T-Large &24 & 1024 & 16 \\
        \bottomrule
    \end{tabular*}

\end{table}
\begin{table*}[!htb]
\caption{\textbf{Comparing~\model~variants} on different pre-training strategies and backbones.
All experiments are trained on Cityscapes train fine set with batch size 8, and evaluated using the single scale test protocol on the Cityscapes validation set in mean IoU (\%) rate.
``Pre'' denotes the pre-training of Transformer part.
``R'' means the Transformer part is randomly initialized.
``w.o.aux'' means ``without auxiliary loss" and ``2step" means 2 upsample layers.
The bolded values highlight the best performance in each respective category.
}
\label{tab:ablation}
\centering
\begin{tabular*}{\textwidth}{@{\extracolsep\fill}lcccccccc}
\toprule
Method & \multicolumn{1}{c}{Pretrain} & Backbone   & \#Params & {FLOPs} & {FPS} & {Mem} & 40k   & 80k   \\
\midrule
FCN~\cite{mmsegmentation}                        & 1K                   & R-101 &   69M   & 620G  & 24 & 1237M &  73.93    &  75.52     \\
Semantic FPN~\cite{mmsegmentation}               & 1K                  & R-101 &  48M & 146G   & 133  & 1074M &   -  &   75.80    \\
\midrule
\naiveModel                & 21K                  & T-Large    & 306M & 960G & 21 & 3321M & 77.37 & 77.90 \\
\mlaModel                   & 21K                  & T-Large    & 311M & 972G & 20 & 3353M & 76.65 & 77.24 \\
\textbf{\pupModel}                   & 21K                  & T-Large    & 318M & 1083G & 18 & 3582M & \textbf{78.39} & \textbf{79.34} \\
\midrule
\pupModel                & R                  & T-Large    & 318M  & 1083G & 18 & 3582M & 42.27 & - \\
\midrule
\smallNaiveModel           & 21K                  & T-Base     & 88M & 296G & 7  &1472M & 75.54 & 76.25 \\
\smallMlaModel                 & 21K                  & T-Base     & 93M & 307G & 7  &1507M & 75.60 & 76.87 \\
\textbf{\smallPupModel}                 & 21K                  & T-Base     & 98M & 417G & 6  & 1877M & \textbf{76.71} & \textbf{78.02} \\
\midrule
\NaiveDeit            & 1K                   & T-Base     & 88M  & 296G &  7 &1472M & 77.85      &   78.66    \\
\MlaDeit              & 1K                   & T-Base     & 93M  & 307G & 7 &1507M & 78.04      &    78.98   \\
\textbf{\PupDeit}              & 1K                   & T-Base     & 98M  & 417G & 6 &1877M &  \textbf{ 78.79}     &    \textbf{ 79.45} \\ 
\midrule
{\MlaBeit}             & {21K}         & {T-Large}    & {311M} & {972G} & {20} & {3353M}    &  {79.59}     &   {80.20}\\
\textbf{\PupBeit}              & {21K}                   & {T-Large}     & {318M} & {1083G} & {18} & {3582M}    &{{\bf 79.84}}     &   {{\bf 80.43}}   \\
\midrule
{\naiveModel\textit{-w.o.aux}}                & {21K}                  & {T-Large}   & {306M} & {960G} & {21} & {3321M} & {76.73} & {77.23} \\
{\pupModel\textit{-2step}}                & {21K}                  & {T-Large}    & {308M} & {990G} & {18} & {3455M} & {77.48} & {78.58} \\
\bottomrule
\end{tabular*}
\end{table*}

\subsection{Single-stage Transformer Setup}
We conduct experiments on three representative semantic segmentation benchmark datasets to evaluate performance on single-stage Transformer.

\noindent \textbf{Semantic segmentation datasets}
\textbf{Cityscapes \cite{Cityscapes}}
densely annotates 19 object categories in images with urban scenes. 
It contains 5000 finely annotated images, split into 2975, ~500 and 1525 for training, validation and testing respectively. 
The images are all captured at a high resolution of $2048 \times 1024$.
In addition, it provides 19,998 coarse annotated images for model training.
\textbf{ADE20K \cite{ADE20K} }
is a challenging scene parsing benchmark with 150 fine-grained semantic concepts.
It contains 20210, ~2000 and 3352 images for training, validation and testing.

\begin{table*}[t]
\caption{\textbf{
Comparison to FCN with different pre-training}
with single-scale inference on the ADE20K val and Cityscapes val set.
The bolded values highlight the best performance in each respective category.
}
\label{tab:pretrain}
\centering
\begin{tabular*}{\textwidth}{@{\extracolsep\fill}lcccccccc}
\toprule

\multirow{2}{*}{Method} & \multirow{2}{*}{Pre} & \multirow{2}{*}{Backbone} & \multicolumn{3}{c}{ADE20K} & \multicolumn{3}{c}{Cityscapes}\\
\cmidrule{4-6}\cmidrule{7-9}
  & & & {FLOPs} & {FPS} & mIoU & {FLOPs}  & {FPS} & mIoU\\
  
\midrule
FCN ~\cite{mmsegmentation} & 1K & R-101 & 276G & 111 & 39.91 & 620G & 24 &73.93 \\
FCN  & 21K & R-101 & 276G & 111 & 42.17 & 620G & 24 & 76.38 \\
\midrule
\mlaModel  & 21K & T-Large  & 368G & 80 & 48.64  & 972G&  7& 76.65  \\
\pupModel & 21K & T-Large   & 426G  & 70 & 48.58  & 1083G& 6& 78.39 \\
\midrule
\MlaDeit & 1K & T-Base & 113G & 204 &46.15  & 307G& 20 &78.98  \\
\PupDeit  & 1K & T-Base &  170G & 181 &46.24  & 417G&  18 & 79.45  \\
\midrule
\MlaBeit  & 21K & T-Large  & 368G & 80 & \textbf{53.0}  & 972G&  7& 80.20  \\
\PupBeit  & 21K & T-Large   & 426G  & 70 & 52.5  & 1083G& 6 & \textbf{80.43} \\

\bottomrule
\end{tabular*}
\end{table*}

\noindent \textbf{Implementation details}
Following the default setting (\eg,~data augmentation and training schedule) of public codebase {\em mmsegmentation} \cite{mmsegmentation},
(i) we apply random resize with ratio between 0.5 and 2, random cropping (768,~512 and 480 for Cityscapes and ADE20K respectively) and random horizontal flipping during training for all the experiments;
(ii) We set batch size 16 and the total iteration to 160,000 and 80,000 for the experiments on ADE20K.
For Cityscapes, we set batch size to 8 with a number of training schedules reported in Table~\ref{tab:ablation},~\ref{tab:city_val} and~\ref{tab:citytest} for fair comparison.
We adopt a polynomial learning rate decay schedule~\cite{pspnet} and employ SGD as the optimizer.
Momentum and weight decay are set to 0.9 and 0 respectively for all the experiments on the three datasets.
We set initial learning rate 0.001 on ADE20K, and 0.01 on Cityscapes.

\noindent \textbf{Auxiliary loss}
As~\cite{pspnet} we also find the auxiliary segmentation loss helps the model training.
Each auxiliary loss head follows a 2-layer network. 
We add auxiliary losses at different Transformer layers:
\naiveModel~($Z^{10}, Z^{15}, Z^{20}$),
\pupModel~($Z^{10}, Z^{15}, Z^{20}, Z^{24}$),
\mlaModel~($Z^6, Z^{12}, Z^{18}, Z^{24}$).
Both auxiliary loss and main loss heads are applied
concurrently.

\begin{figure}[t]\centering
\includegraphics[width=1.0\linewidth]{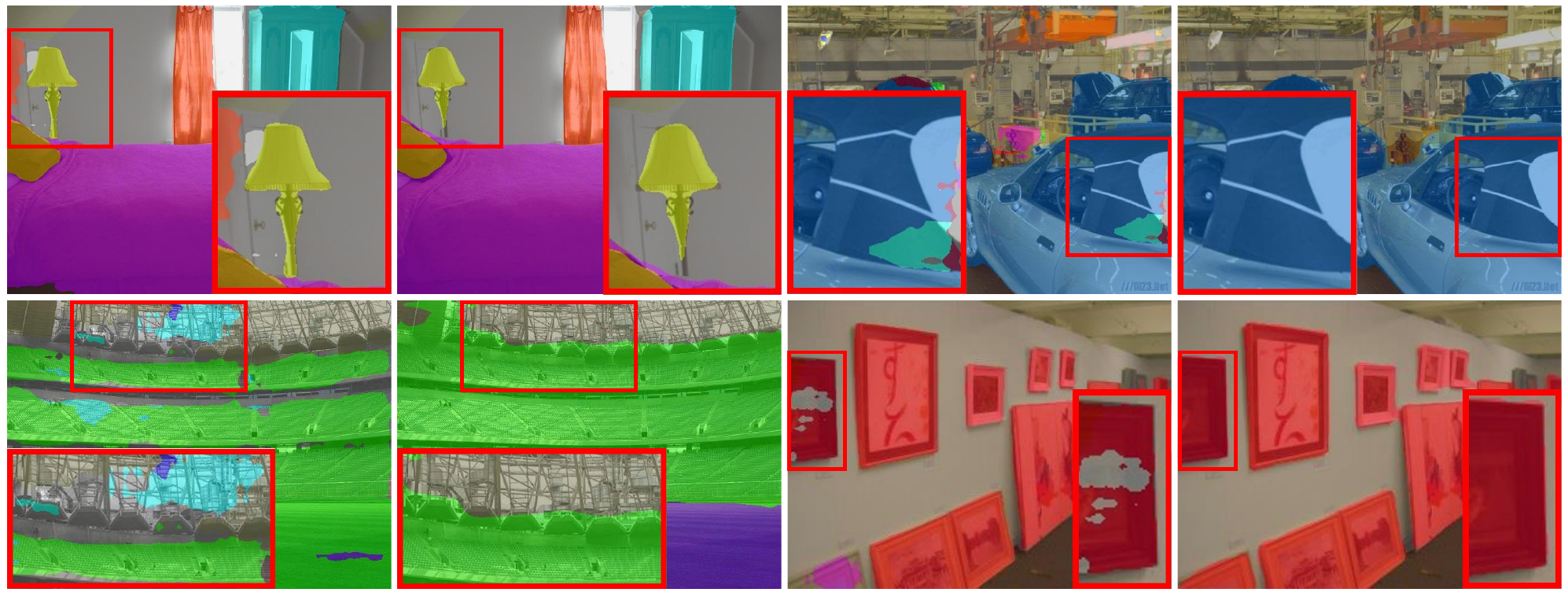}
\caption{\textbf{Qualitative results on ADE20K:} SETR (right column) vs. dilated FCN baseline (left column) in each pair. 
Best viewed in color and zoom in.}
\label{fig:ade}
\end{figure}
\begin{table*}[t]
\caption{\textbf{State-of-the-art comparison on the ADE20K dataset.}
Performances of different model variants
are reported.
SS: Single-scale inference. MS: Multi-scale inference.
The bolded values highlight the best performance in each respective category.}
\label{tab:ade}
\begin{tabular*}{\textwidth}{@{\extracolsep\fill}lccccc}
\toprule
Method &Pre & Backbone & \#Params & Flops & mIoU \\
\midrule
FCN (160k, SS)~\cite{mmsegmentation}&1K & ResNet-101 &69M & 276G  & 39.9 \\
FCN (160k, MS)~\cite{mmsegmentation}&1K & ResNet-101 &69M  & 276G  & 41.4 \\
\midrule
PSPNet~\cite{pspnet} &1K & ResNet-101 & 68M  & 256G & 44.4 \\
 DLabV3+~\cite{deeplabv3p} &1K & ResNet-101 & 62M & 262G & 46.9 \\
 CCNet~\cite{huang2018ccnet} &1K & ResNet-101 & 68M & 278G & 43.7\\

 UperNet~\cite{xiao2018unified} (160k, MS) &1K & Swin-T & 60M & 236G & 45.8\\
 UperNet~\cite{xiao2018unified} (160k, MS) &1K & Swin-S & 81M & 259G & 49.5\\
 UperNet~\cite{xiao2018unified} (160k, MS) &1K & Swin-B & 121M & 297G & 49.7\\
\midrule
\naiveModel~(160k, SS) &21K & ViT-Large &306M & 363G  & 48.1 \\
\naiveModel~(160k, MS) &21K & ViT-Large &306M & 363G & 48.8\\
\pupModel~(160k, SS) &21K & ViT-Large&318M & 426G & 48.6  \\
\pupModel~(160k, MS)  &21K & ViT-Large &318M & 426G & 50.1 \\
\mlaModel~(160k, SS) &21K & ViT-Large &311M & 368G & 48.6  \\
\textbf{\mlaModel~(160k, MS)} &21K & ViT-Large &311M & 368G & \textbf{50.3} \\
\midrule
{\PupBeit ~(160k, SS)}& {21K} & {ViT-Large} & {318M}  & 426G   & {52.5}   \\
{\PupBeit ~(160k, MS)}& {21K} & {ViT-Large} & {318M}    & 426G & {52.7}  \\
{\MlaBeit ~(160k, SS)} & {21K} & {ViT-Large} & {311M} & 318G  & {53.0} \\
\textbf{\MlaBeit ~(160k, MS)} & {21K} & {ViT-Large} & {311M} & 318G & \textbf{53.1} \\
\midrule
\PupDeit ~(160k, SS) &1K & ViT-Base &98M & 170G & 46.3  \\
\PupDeit ~(160k, MS) &1K & ViT-Base &98M & 170G &47.3      \\
\MlaDeit ~(160k, SS) &1K & ViT-Base &93M & 113G &46.2     \\
\textbf{\MlaDeit ~(160k, MS)} &1K & ViT-Base &93M & 113G &\textbf{47.7}    \\
\midrule
{SETR-\textit{PUP-HLG-small} ~(160k, MS)} & {1K} & {HLG-small} & {37M} & {243G} & {47.3}\\
{SETR-\textit{PUP-HLG-Med }~(160k, MS)} & {1K} & {HLG-Med} & {61M} & {269G} & {49.3} \\
\textbf{SETR-\textit{PUP-HLG-Large} ~(160k, MS)} & {1K} & {HLG-Large} & {100M} & {294G} & \textbf{49.8}\\
\bottomrule
\end{tabular*}
\end{table*}
\noindent \textbf{Multi-scale test}
\label{sec:multiscale}
We use the default settings of {\em mmsegmentation} \cite{mmsegmentation}.
Specifically, the input image is first scaled to a uniform size.
Multi-scale scaling and random horizontal flip are then performed on the image with a scaling factor (0.5, 0.75, 1.0, 1.25, 1.5, 1.75). 
Sliding window is adopted for test (\eg, $512 \times 512$ for ADE20K).
If the shorter side is smaller than the size of the sliding window, the image is scaled with its shorter side to the size of the sliding window (\eg, 512) while keeping the aspect ratio.
Synchronized BN is used in decoder and auxiliary loss heads.
For training simplicity, we do not adopt the widely-used tricks such as OHEM~\cite{ocnet} loss in model training.

\noindent \textbf{Baselines}
We adopt dilated FCN~\cite{fcn} and Semantic FPN~\cite{kirillov2019panoptic} as baselines with their results taken from \cite{mmsegmentation}.
Our models and the baselines are trained and tested in the same settings for fair comparison.
In addition, state-of-the-art models are also compared. 
Note that the dilated FCN is with output stride 8 and we use output stride 16 in all our models due to GPU memory constrain.

\noindent \textbf{SETR variants}

Three variants of our model with different decoder designs (see Sec.~\ref{sec:decoder}), namely \naiveModel, \pupModel~  and \mlaModel.
Besides,
we use two variants of the encoder ``T-Base'' and ``T-Large'' with 12 and 24 layers respectively (Table~\ref{tab:config}).
Unless otherwise specified, we use ``T-Large'' as the encoder for \naiveModel, \pupModel~and \mlaModel.
We denote \smallNaiveModel~as the model utilizing ``T-Base'' in \naiveModel.

\noindent \textbf{Pre-training}
We use the pre-trained weights provided by ViT~\cite{dosovitskiy2020image},  DeiT~\cite{touvron2020training} {or 
BEiT~\cite{bao2021beit}}
to initialize all the Transformer layers and the input linear projection layer in our model.
We denote~\NaiveDeit~as the model utilizing DeiT~\cite{touvron2020training} pre-training in \smallNaiveModel{} {and \PupBeit{} as the model model utilizing BEiT~\cite{bao2021beit} pre-training in \pupModel}.
All the layers without pre-training are randomly initialized.
%
For the Transformer part, we use the weights pre-trained by ViT~\cite{dosovitskiy2020image}, DeiT~\cite{touvron2020training}, {BEiT~\cite{bao2021beit}} or randomly initialized.

\begin{figure*}[t]\centering
\includegraphics[width=1.0\linewidth]{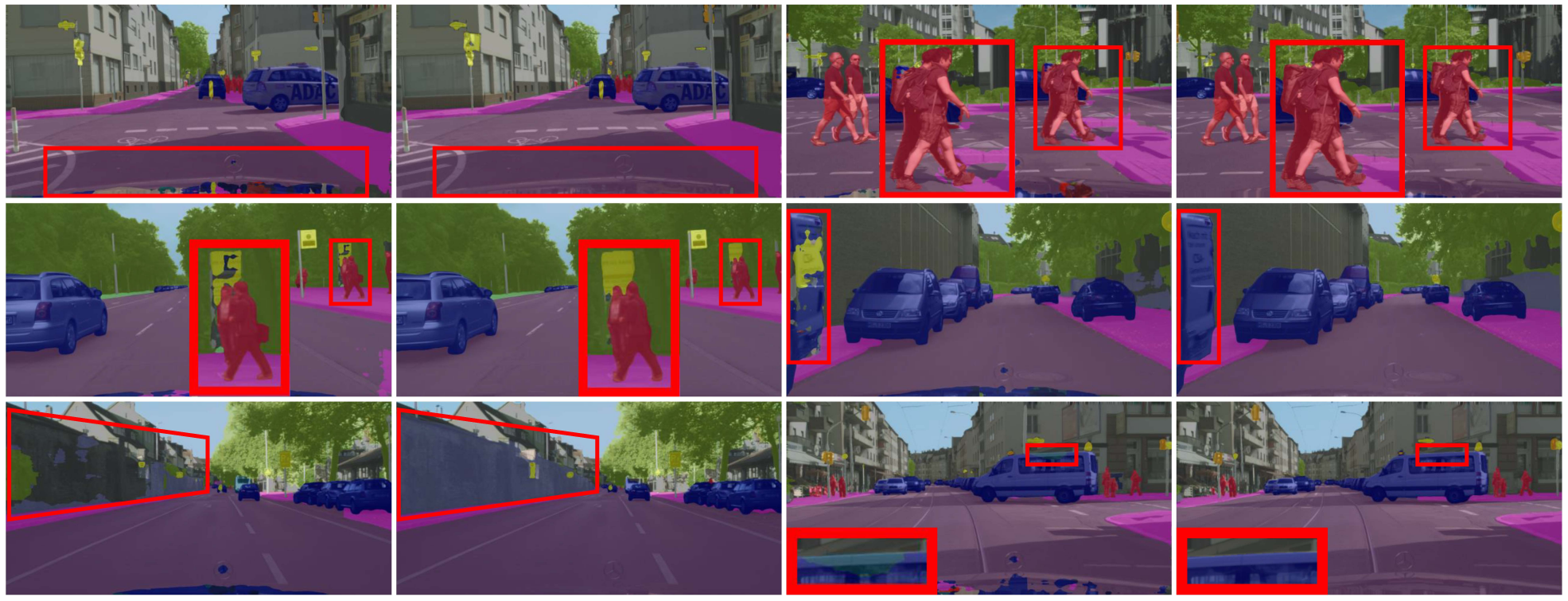}
\caption{\textbf{Qualitative results on Cityscapes:} SETR (right column) vs. dilated FCN baseline (left column) in each pair. 
Best viewed in color and zoom in.}
\label{fig:city}
\end{figure*}
\begin{table*}[t]
\caption{\textbf{State-of-the-art comparison on the Cityscapes validation set.}
Performances of different training schedules (\eg, 40k and 80k) are reported.
SS: Single-scale inference. MS: Multi-scale inference.
The bolded values highlight the best performance in each respective category.}
\label{tab:city_val}

\begin{tabular*}{\textwidth}{@{\extracolsep\fill}lccccc}
\toprule
Method  & Pre & Backbone & \#Params & Flops & mIoU \\
\midrule
FCN (40k, SS)~\cite{mmsegmentation} & 1K  & ResNet-101 & 68M  & 619G & 73.9 \\
FCN (40k, MS)~\cite{mmsegmentation} & 1K  & ResNet-101 & 68M  & 619G & 75.1 \\
FCN (80k, SS)~\cite{mmsegmentation} & 1K  & ResNet-101 & 68M  & 619G & 75.5 \\
FCN (80k, MS)~\cite{mmsegmentation} & 1K  & ResNet-101 & 68M  & 619G & 76.6 \\
\midrule
PSPNet~\cite{pspnet} & 1K  & ResNet-101  & 68M  & 576G & 78.5 \\
DLabV3+~\cite{deeplabv3p} & 1K & ResNet-101 & 62M & 571G & 79.3  \\
NonLocal~\cite{wang2018nonlocal} & 1K  & ResNet-101 & 66M & 669G & 79.10 \\
CCNet~\cite{huang2018ccnet}  & 1K  & ResNet-101 & 68M & 625G  & 80.2 \\
GCNet~\cite{cao2019gcnet} & 1K  & ResNet-101 & 66M & 620G & 78.1\\
OCRNet~\cite{yuan2019object} & 1K & HRNet-W48 & 70M & 972G & 81.1 \\
\midrule
\mlaModel~(80k, MS) & 21K & ViT-Large & 311M & 972G & 79.1 \\
\textbf{\pupModel~(80k, MS)} & 21K & ViT-Large & 318M & 1083G & \textbf{82.2}  \\
\midrule
\MlaDeit~(80k, MS) & 1K  & ViT-Base & 93M & 307G & 78.9\\
\textbf{\PupDeit~(80k, MS)} & 1K & ViT-Base & 98M & 417G & \textbf{79.5}  \\
\midrule
{SETR-\textit{PUP-HLG-small} ~(80k, MS)} & {1K} & {HLG-small} & {37M} & {556G} & {81.8}\\
{SETR-\textit{PUP-HLG-Med }~(80k, MS)} & {1K} & {HLG-Med} & {61M} & {619G} & {82.5}\\
\textbf{SETR-\textit{PUP-HLG-Large} ~(80k, MS)} & {1K} & {HLG-Large} & {100M} & {791G} & {\textbf{82.9}}\\
\bottomrule
\end{tabular*}
\end{table*}

We use patch size $16\times16$ for all the experiments. We perform 2D interpolation on the pre-trained position embeddings, according to their location in the original image for different input size fine-tuning.

\noindent \textbf{Evaluation metric}
Following the standard evaluation protocol~\cite{Cityscapes}, the metric of mean Intersection over Union (mIoU) averaged over all classes is reported.
For ADE20K, additionally pixel-wise accuracy is reported following the existing practice.

\subsection{HLG Transformer Setup}
In this section, we focus on the evaluation of the proposed HLG Transformers.
We first evaluate the image classification task using our HLG Transformers as the backbone. 
For more extensive and task agnostic evaluations, we further test object detection, instance segmentation, and semantic segmentation.

\noindent \textbf{Image classification dataset} 
The ImageNet-1K dataset \cite{imagenet} contains 1.28 million training images and 50K validation images from 1,000 object categories used for model training and evaluation respectively.

\noindent \textbf{Image classification details}

We validate the performance of a HLG Transformer by training on the training set and reporting the Top-1 accuracy on the validation set. For a fair comparison, we follow the setting of Swin Transformer \cite{liu2021swin} and DeiT \cite{touvron2020training}. For data augmentation, we perform random-size cropping to $ 224 \times 224 $, and adopt random horizontal flipping \cite{googlenet} and mix up \cite{zhang2017mixup}. 
Training from scratch, we set the batch size to 1024 and the epochs to 300. We use an initialization learning rate of 0.001, a cosine decay learning rate scheduler with 20 epochs of linear warm-up. We employ the AdamW optimizer with a weight decay 0f 0.05. To prevent model overfitting, the drop-path \cite{larsson2016fractalnet} rate is set to 0.1 by default and 0.3 for HLG-Large due to its stronger capacity.

\noindent \textbf{Semantic segmentation details}
We use the segmentation framework as SETR-PUP.
For more extensive test, we also evaluate UperNet \cite{xiao2018unified} as a second framework.
Each backbone network is pretrained on ImageNet-1k. For a fair comparison, we use the same data augmentation and training schedule as the Swin Transformers \cite{liu2021swin}. The training images are randomly deformed at a ratio of 0.5$\sim$2 with a cropped size of $ 768 \times 768 $ for Cityscapes. Random horizontal flip operation is performed during training. The number of training iteration is set to 40K for Cityscapes. The optimizer uses stochastic gradient descent with momentum of 0.9, weight decay of 0 and batch size of 8. We adopt the poly learning rate adjustment strategy with the initial learning rate 0.01 for Cityscapes.
Only single-scale inference introduced is used.

\noindent \textbf{Object detection dataset} 

For object detection experiments, we use the COCO 2017 dataset with 118K training and 5K validation images \cite{COCO_dataset}.

\noindent \textbf{Object detection details}

For object detection and instance segmentation,
we take RetinaNet \cite{lin2017focal} and Mask R-cnn \cite{he2017mask} as the detection framework.
The backbone (\eg, ResNet and our HLG Transformer) is initialized with the weights pre-trained on the ImageNet-1k dataset, with the rest initialized by Xavier. We follow the common training settings. We set the batch size to 16 with 1x training schedule (\ie, 12 epochs). We use the AdamW optimizer with a weight decay of $ 1 \times 10^{-4} $ and initial learning rate of $ 1 \times 10^{-4} $. Multi-scale training is adopted, while the training images are randomly resized to the shorter side not exceeding 800 pixels and the longer side not exceeding 1333 pixels.

\subsection{Vision Transformer Pretraining}
\begin{table*}[t]
\caption{Vision Transformer pretraining results on the ILSVRC-2012 ImageNet validation dataset.
Performances of different model variants
are reported.
The bolded values highlight the best performance in each respective category.
}
\label{tab:hlg_classification}
\begin{tabular*}{\textwidth}{@{\extracolsep\fill}lcccc}
\toprule
 Method & Image size & \#Params & FLOPs & Top1(\%) \\
 \midrule
 ViT-Large-1K~\cite{dosovitskiy2020image} & $384^2$ & 307M & 44.3G & 76.5 \\
 ViT-Large-21K~\cite{dosovitskiy2020image} & $384^2$ & 307M & 44.3G & 85.2 \\
 BeiT-B-21K~\cite{bao2021beit} & $384^2$ & 307M & 44.3G & 86.3\\
 DeiT-B-1K~\cite{touvron2022deit} & $384^2$ & 87M & 49.6G &  83.1\\
\midrule

 DeiT-Ti~\cite{touvron2020training} & $224^2$ & 5.7M & 1.3G & 72.2 \\
 CrossViT-Ti~\cite{chen2021crossvit} & $224^2$ & 6.9M & 1.6G & 73.4 \\
\textbf{HLG-Mobile}& $224^2$  & {4.3M} & {0.9G} & {\textbf{75.1}} \\
\midrule

 PiT-XS ~\cite{heo2021rethinking}& $224^2$  & 10.6M & 1.4G & 78.1  \\
 ConT-S ~\cite{yan2021contnet}& $224^2$  & 10.1M & 1.5G & 76.5  \\
 PVT-T ~\cite{wang2021pyramid}& $224^2$  & 13.2M & 1.9G & 75.1  \\
 ConViT-Ti+ ~\cite{d2021convit}& $224^2$  & 10.0M & 2.0G & 76.7  \\
 ViP-Ti ~\cite{bai2021visual}& $224^2$  & 12.8M & 1.7G & 79.0  \\
\textbf{HLG-Tiny}& $224^2$  & {11.0M} & {2.1G} & {\textbf{81.1}}  \\
 \midrule

 LocalViT-S ~\cite{li2021localvit}& $224^2$  & 22.4M & 4.6G & 80.8  \\
 ConViT-S ~\cite{d2021convit}& $224^2$  & 27.0M & 5.4G & 81.3  \\
 NesT-S ~\cite{zhang2021aggregating}& $224^2$  & 17.0M & 5.8G & 81.5  \\
 Swin-T ~\cite{liu2021swin}& $224^2$  & 29.0M & 4.5G & 81.3  \\
 CoAtNet-0~\cite{dai2021coatnet}& $224^2$  & 25.0M & 4.2G & 81.6\\
 DeiT III-S~\cite{touvron2022deit}& $224^2$  & 22.0M & 4.6G & 81.4\\
SwinV2-T~\cite{liu2022swin}& $256^2$  & 29.0M & 4.5G & 81.7  \\
ConvNext-T~\cite{liu2022convnet}& $224^2$  & 29.0M & 4.5G & 82.1  \\
\textbf{HLG-Small}& $224^2$  & {24.2M} & {4.7G} & {\textbf{82.3}}    \\
 \midrule
 ConT-M ~\cite{yan2021contnet}& $224^2$  & 39.6M & 6.4G & 81.8  \\
 Twins-PCPVT-B ~\cite{chu2021twins}& $224^2$  & 43.8M & 6.4G & 82.7  \\
 PVT-M ~\cite{wang2021pyramid}& $224^2$  & 44.2M & 6.7G & 81.2  \\
 Swin-S ~\cite{liu2021swin}& $224^2$  & 50.0M & 8.7G & 83.0  \\
ConvNext-S ~\cite{liu2022convnet}& $224^2$  & 50.0M & 8.7G & 83.1  \\
 CoAtNet-1~\cite{dai2021coatnet}& $224^2$  & 42.0M & 8.4G & 83.3\\
SwinV2-S ~\cite{liu2022swin}& $256^2$  & 50.0M & 8.7G & 83.6  \\
\textbf{HLG-Medium}& $224^2$  & {43.7M} & {9.0G} & {\textbf{83.6}}   \\
 \midrule
 DeiT-B ~\cite{touvron2020training}& $224^2$  & 86.6M & 17.6G & 81.8  \\
 PiT-B ~\cite{heo2021rethinking}& $224^2$   & 73.8M & 12.5G & 82.0  \\
 ConViT-B ~\cite{d2021convit}& $224^2$  & 86.0M & 17.0G & 82.4  \\
 Swin-B ~\cite{liu2021swin}& $224^2$  & 88.0M & 15.4G & 83.3 \\
 DeiT III-B~\cite{touvron2022deit}& $224^2$  & 86.6M & 15.5G & 83.8\\
 ConvNext-S ~\cite{liu2022convnet}& $224^2$  & 89.0M & 15.4G & 83.8  \\
 CoAtNet-2~\cite{dai2021coatnet}& $224^2$  & 75.0M & 15.7G & 84.1\\
SwinV2-B ~\cite{liu2022swin}& $256^2$  & 88.0M & 15.4G & 84.1 \\
\textbf{HLG-Large}& $224^2$  &  {84.2M} & {15.9G} & {\textbf{84.1}}  \\
\bottomrule
\end{tabular*}
\end{table*}
\begin{table*}[t]
\caption{{Throughput, peek memory and image classification results on ImageNet-1K V2~\cite{recht2019imagenet} matched frequency dataset.
The bolded values highlight the best performance in each respective category.
}}
\label{tab:hlg_imagenetv2}
\begin{tabular*}{\textwidth}{@{\extracolsep\fill}lcccccc}
 \toprule
  Method & Size & \#Params & FLOPs & FPS & Memory & Top1 (\%) \\
\midrule
{Swin-T} & {$224^2$} & {29M} & {4.5G} & {755} & {11.6GB} & {69.7}\\
{Swin-S} & {$224^2$} & {50M} & {8.7G} & {437} & {18.6GB} & {72.1}\\
{Swin-B} & {$224^2$} & {88M} & {15.4G} & {278} & {25.2GB} & {72.3}\\
\midrule
{HLG-Tiny} & {$224^2$} & {11M} & {2.1G} & {734} & {11.7GB} & {70.1}\\
{HLG-Small} & {$224^2$} & {24M} & {4.7G} & {697} & {17.5GB} & {70.3}\\
{HLG-Medium} & {$224^2$} & {44M} & {9.0G} & {392} & {23.6GB} & {72.4}\\
\textbf{HLG-Large} & {$224^2$} & {84M} & {15.9G} & {221} & {31.6GB} & {\textbf{73.2}}\\
\bottomrule
\end{tabular*}
\end{table*}
{We initially examine the pretraining of Vision Transformers. 
Table~\ref{tab:hlg_classification} compares the results of HLG Transformer with current state-of-the-art ViTs on ImageNet-1k and 21k. 
The first part of the table showcases the pretrained single-stage Vision Transformers: ViT~\cite{dosovitskiy2020image}, BeiT~\cite{bao2021beit}, and DeiT~\cite{touvron2022deit} we employ. 
It's evident that single-stage vision transformers are heavily reliant on data scale, with a substantial difference observed between 21K and 1K pretrained ViTs.
}

{
Therefore, the hierarchical Vision Transformer addresses the computation cost associated with single-stage Vision Transformer pretraining, allowing for the use of less training data while achieving comparable results.
For the hierachial Vision Transformer, we make the following observations.
(1) Compared to the DeiT models, our proposed HLG Transformer
is clearly superior in both accuracy and efficiency.
For example, HLG-Mobile/Large outperforms DeiT-Ti/B by 2.2\%/1.6\% in accuracy whilst enjoying less parameters and FLOPs.
This is attributed to the hierarchical architecture design
and our HLG Transformer design.
(2) Compared with the PVT models, all HLG variants remain significantly better.
For instance, HLG-Tiny with 11.0M parameters achieves 81.1\% accuracy
in comparison 75.1\% by PVT-T at similar size. 
This validates the superiority of our HLG Transformer building block
over PVT's linear attention.
(3) Compared with the Swin models similarly characterized by local attentive representation learning, our HLG Transformer counterparts are still more effective in most cases.
Particularly, in low-consumption regime, HLG-Small with 34.2M parameters achieves 82.3\% accuracy, yielding a margin of 1.0\% over Swin-T. For larger models, HLG Transformer achieves a better trade-off between accuracy and model size. Swin-B with 88.0M parameters reaches 83.3\%, while our HLG-Large achieves 84.1\% with 3.8M fewer parameters.
}

In Table~\ref{tab:hlg_imagenetv2} we show the superiority of our models
in the trade-off among efficiency, memory and accuracy on ImageNet1K-V2~\cite{recht2019imagenet} matched frequency dataset, without model overfitting. 

\subsection{Results of Semantic Segmentation}
\noindent \textbf{Ablation Studies}
Table~\ref{tab:ablation} and~\ref{tab:pretrain} show ablation studies on 
{\bf(a)} different variants of \model~on various training schedules,
{\bf(b)} comparison to FCN~\cite{mmsegmentation} and Semantic FPN~\cite{mmsegmentation},
{\bf(c)} pre-training on different data,
and {\bf(d)} comparison to FCN with different pre-training.
Unless specified otherwise, all experiments in Table~\ref{tab:ablation} and~\ref{tab:pretrain} are trained on Cityscapes train fine set with batch size 8, and evaluated using the single scale test protocol on the Cityscapes validation set in mean IoU (\%) rate.
Experiments on ADE20K also follow the single scale test protocol.
\begin{table*}[t]
\caption{\textbf{Comparison on the Cityscapes test set.}
$\ddag$: trained on fine and coarse annotated data.
The bolded values highlight the best performance in each respective category.
}
\label{tab:citytest}
\begin{tabular*}{\textwidth}{@{\extracolsep\fill}llc}
\toprule
Method & Backbone & mIoU\\
\midrule
PSPNet~\cite{pspnet} & ResNet-101 & 78.40 \\
DenseASPP~\cite{denseaspp} & DenseNet-161 & 80.60 \\
BiSeNet~\cite{bisenet} & ResNet-101 & 78.90 \\
PSANet~\cite{psanet} & ResNet-101 & 80.10 \\
DANet~\cite{DAnet} & ResNet-101 & 81.50 \\
OCNet~\cite{ocnet} & ResNet-101 & 80.10 \\
CCNet~\cite{huang2018ccnet} & ResNet-101 & 81.90 \\
Axial-DeepLab-L~\cite{wang2020axial} & Axial-ResNet-L & 79.50 \\
Axial-DeepLab-XL~\cite{wang2020axial} & Axial-ResNet-XL & 79.90 \\
\midrule
\pupModel~(100k) & T-Large & 81.08\\
\textbf{\pupModel}$^\ddag$ & T-Large & \textbf{81.64} \\
\bottomrule
\end{tabular*}
\end{table*}
{FPS and memory are tested by single GPU during inference.}

\begin{figure}[t]\centering
\includegraphics[width=1.02\linewidth]{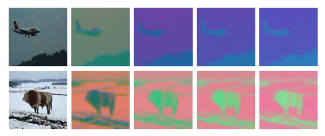}
\vspace{-0.3cm}
\caption{Visualization of output feature of layer $Z^1$, $Z^9$, $Z^{17}$, $Z^{24}$ of SETR trained on Pascal Context.
Best viewed in color.}
\label{fig:layer}
\end{figure}

From Table~\ref{tab:ablation}, we can make the following observations: 
{\bf(i)} Progressively upsampling the feature maps, \pupModel~achieves the best performance among all the variants on Cityscapes.
One possible reason for inferior performance of \mlaModel~is that the feature outputs of different Transformer layers do not have the benefits of resolution pyramid as in feature pyramid network (FPN) (see Figure~\ref{fig:layer}).
However, \mlaModel~performs slightly better than \pupModel, and much superior to the variant \naiveModel~that upsamples the Transformers output feature by 16$\times$ in one-shot, on ADE20K val set (Table~\ref{tab:pretrain} and~\ref{tab:ade}).
{\bf(ii)} The variants utilizing "T-Large" (e.g., \mlaModel and \naiveModel) outperform their "T-Base" counterparts, namely \smallMlaModel and \smallNaiveModel, as anticipated. {However, employing superior pre-training techniques on "T-Base" such as DeiT~\cite{touvron2022deit} yields even better performance for these smaller Vision Transformer models on Cityscapes compared to their larger counterparts. This underscores the significant impact of pre-training methods on performance.}
{\bf(iii)} Pre-training is critical for our model.
Randomly initialized~\pupModel~only gives 42.27\% mIoU on Cityscapes.
{
Model pre-trained with BEiT~\cite{bao2021beit} on ImageNet-21K gives the best performance on Cityscapes and ADE20K, surpassing the alternative pre-trained with ViT~\cite{dosovitskiy2020image} on ImageNet-21K.
Additionally, employing distillation, Deit~\cite{touvron2022deit} pretrained on ImageNet-1K and "T-Base" Vision Transformer perform better than ImageNet-1K pretrained "T-Large" Vision Transformer on Cityscapes but worse on ADE20K. This indicates that the reliance of ADE20K on dataset scale for pretraining impacts its performance more significantly due to its diverse range of variants and ample datasets.
With appropriate pre-training techniques, employing a smaller model and a smaller pretraining dataset can outperform larger models pre-trained on larger datasets.
}
{\bf(iv)} 
To study the power of pre-training and further verify the effectiveness of our proposed approach, we conduct the ablation study on the pre-training strategy.
For fair comparison with the FCN baseline,
we first pre-train a ResNet-101 on the ImageNet-21k dataset with a classification task and then adopt the pre-trained weights for a dilated FCN training for semantic segmentation on the target dataset (ADE20K or Cityscapes).
As shown in Table~\ref{tab:pretrain}, with ImageNet-21k pre-training the FCN baseline experiences a clear improvement over the variant pre-trained on ImageNet-1k.
However, our method outperforms the FCN counterparts by a large margin, verifying that the advantage largely comes from the proposed \textit{sequence-to-sequence} representation learning strategy rather than bigger pre-training data.
{{\bf(v)} 
The study on auxiliary loss in Table~\ref{tab:ablation} demonstrates the significant impact of auxiliary loss. It notably enhances model performance, particularly evident at 40,000 iterations. The auxiliary loss aids in regulating the features of the Vision Transformer in early layers, thereby accelerating optimization.
}
{{\bf(vi)} 
We also investigate the decoder head of \pupModel. We vary the number of upsample convolution layers from 4 to 2 and adjust the upsampling factor to $4\times$. The results indicate a 0.7\% reduction compared to the 4-layer PUP. This suggests that maintaining more upsampling layers preserves accuracy during the upsampling operation.
}
{{\bf(vii)} 
We compare the FLOPS, inference speed, and GPU memory between FCN and different variants of Vision Transformer. As expected, Vision Transformer incurs higher FLOPS, time, and GPU memory usage compared to convolutional models. The increased FLOPS primarily arise from the MLP and self-attention mechanisms in Transformer. Additionally, the larger memory usage mainly stems from the large matrices involved in the scale product of self-attention, which incurs a memory complexity of $\mathcal{O}(n^2)$.
Despite the higher computational cost, Vision Transformer boasts overwhelming pretraining scalability. With access to more data and improved pretraining techniques, Vision Transformer-based segmentation models can achieve superior results compared to the top-performing convolutional models. Moreover, as research advances in higher-performance computation~\cite{dao2022flashattention}, the cost of Transformers is expected to witness significant optimization.
}

\noindent \textbf{Results on ADE20K}
{
Table~\ref{tab:ade} presents our results on the challenging ADE20K dataset.
Our \mlaModel~achieves superior mIoU of 48.64\% with single-scale (SS) inference.
When multi-scale inference is adopted, our method achieves a new state of the art with mIoU hitting 50.3\%.
Utilizing the same model, the latest pretraining technique, ViT: BeiT~\cite{bao2021beit}, elevates the SETR to the state of the art with an mIoU of 53.1
Figure~\ref{fig:ade} shows the qualitative results of our model and dilated FCN on ADE20K.
When training a single model on the train+validation set with the default 160,000 iterations, our method ranks $1^{st}$
place in the highly competitive ADE20K test server leaderboard.
}

{
However, single-stage Transformers suffer from high computational and pretraining costs. Even with DeiT~\cite{touvron2022deit} to reduce computation costs, the results of SETR paired with DeiT show a significant decline. Thus, we introduce the cost-efficient backbone, Hierarchical Local Global Transformer (HLG).
Under similar computational costs, HLG equipped with SETR-PUP achieves the best results in the ImageNet 1K pretraining setting and yields comparable results to SETR-PUP equipped with ViT Large. Additionally, it outperforms the Swin Transformer with UperNet as the decoder head across all parameter levels.
}

\noindent \textbf{Results on Cityscapes}
{
Tables~\ref{tab:city_val} and~\ref{tab:citytest} show the comparative results on the validation and test set of  Cityscapes respectively. 
We can see that our model \pupModel~is superior to FCN baselines, and FCN plus attention based approaches, such as Non-local~\cite{wang2018nonlocal} and CCNet~\cite{huang2018ccnet}; and its performance is on par with the best results reported so far. 
Similar to the observation in the ADE20K dataset, reducing ViT-Large to ViT-Base results in a significant drop in performance. 
Therefore, after applying the Hierarchical Local Global Transformer, SETR achieves the best performance with significantly fewer parameters and less computational overhead.
Figure~\ref{fig:city} shows the qualitative results of our model and dilated FCN on Cityscapes.}

\subsection{Results of More Dense Prediction Tasks}

\noindent \textbf{Object Detection}
\begin{table*}[t]
\caption{Object detection performance using our HLG Transformer with RetinaNet on COCO validation set.
The bolded values highlight the best performance in each respective category.}
\label{tab:hlg_detection_retinanet}
\begin{tabular*}{\textwidth}{@{\extracolsep\fill}lccccccc}
\toprule
Method & \#Params 
& $ AP $ & $ AP_{50} $  & $ AP_{75} $ & $ AP_{S} $ & $ AP_{M} $  & $ AP_{L} $ \\
\midrule
ResNet18 ~\cite{resnet} & 21.3M & 31.8  & 49.6  & 33.6  & 16.3  & 34.3  & 43.2 \\
PVT-T ~\cite{wang2021pyramid} & 23.0M & 36.7 & 56.9 & 38.9 & 22.6 & 38.8 & 50.0 \\
PVTv2-B1 ~\cite{wang2021pvtv2} & 23.8M & 41.2 & 61.9 & 43.9 & 25.4 & 44.5 & 54.3 \\
\textbf{HLG-Tiny} & {20.7M} & {\textbf{43.3}} & {64.4} & {46.1} & {28.7} & {47.6} & {56.9} \\
\midrule
ResNet50 ~\cite{resnet} & 37.7M & 36.3 & 55.3 & 38.6 & 19.3 & 40.0 & 48.8  \\
ConT-M ~\cite{yan2021contnet} & 27.0M & 39.3 & 59.3 & 41.8 & 23.1 & 43.1 & 51.9 \\
PVT-S ~\cite{wang2021pyramid} & 34.2M & 40.4 & 61.3 & 43.0 & 25.0 & 42.9 & 55.7  \\
ViL-S ~\cite{zhang2021multi} & 35.7M & 41.6 & 62.5 & 44.1 & 24.9 & 44.6 & 56.2  \\
Swin-T ~\cite{liu2021swin} & 38.5M & 41.5 & 62.1 & 44.2 & 25.1 & 44.9 & 55.5  \\
RegionViT-S+ ~\cite{chen2021regionvit} & 41.6M & 43.9 & 65.5 & 47.3 & 28.5 & 47.3 & 57.9  \\
\textbf{HLG-Small} & {34.4M}  & {\textbf{44.4}} & {65.6} & {47.8} & {30.0} & {48.6} & {57.9}  \\
\midrule
ResNet101 ~\cite{resnet} & 56.7M & 38.5  & 57.8 & 41.2  & 21.4  & 42.6  & 51.1 \\
PVT-M ~\cite{wang2021pyramid} & 53.9M & 41.9 & 63.1 & 44.3 & 25.0 & 44.9 & 57.6 \\
ViL-M ~\cite{zhang2021multi} & 50.8M & 42.9 & 64.0 & 45.4 & 27.0 & 46.1 & 57.2 \\
ViP-M ~\cite{bai2021visual} & 48.8M & 44.3 & 65.9 & 47.4 & 30.7 & 48.0 & 57.9 \\
Swin-S ~\cite{liu2021swin} & 59.8M & 44.5 & 65.7 & 47.5 & 27.4 & 48.0 & 59.9 \\
\textbf{HLG-Medium} & {57.9M} & {\textbf{46.6}} & {67.9} & {50.2} & {31.4} & {51.0} & {60.7} \\
\midrule
ResNeXt101 ~\cite{xie2017aggregated} & 95.5M & 41.0 & 60.9 & 44.0 & 23.9 & 45.2 & 54.0 \\
PVT-L ~\cite{wang2021pyramid} & 71.1M & 42.6 & 63.7 & 45.4 & 25.8 & 46.0 & 58.4 \\
ViL-B ~\cite{zhang2021multi} & 66.7M & 44.3 & 65.5 & 47.1 & 28.9 & 47.9 & 58.3 \\
Swin-B ~\cite{liu2021swin} & 98.4M & 44.7 & 65.9 & 49.2 & - & - & - \\
RegionViT-B+ ~\cite{chen2021regionvit} & 84.5M & 46.1 & 68.0 & 49.5 & 30.5 & 49.9 & 60.1 \\
\textbf{HLG-Large} & {94.8M} & {\textbf{47.7}} & {69.0} & {51.2} & {32.8} & {51.7} & {62.2} \\
\bottomrule
\end{tabular*}
\end{table*}

Table~\ref{tab:hlg_detection_retinanet} shows the bounding box AP results of the RetinaNet detection framework \cite{lin2017focal} in different backbones, including both CNNs and ViTs. 
We have several observations.
(1) Compared with top CNNs (\eg, ResNet and ResNeXt), 
ViTs with the similar parameters can improve the average precision by up to 10.7\%.
(2) Among ViTs, it is evident that our HLG Transformer achieves the best results across all the size groups.
Specifically, compared with PVTv2-B1, HLG-Tiny use a smaller number of parameters to reach a margin of 1.3\%. 
HLG-Large improves over Swin-B and RegionViT-B+ by 3.0\% and 1.6\% in AP, respectively.

\begin{table*}[t]
\caption{Instance segmentation performance using our HLG Transformer with Mask R-CNN on COCO validation set.
$ AP^b $ denotes bounding box AP and $ AP^m $ denotes mask AP.
The bolded values highlight the best performance in each respective category.}
\label{tab:hlg_detection_maskrcnn}
\begin{tabular*}{\textwidth}{@{\extracolsep\fill}lccccccc}
\toprule
Method & \#Params
& $ AP^b $ & $ AP^{b}_{50} $  & $ AP^{b}_{75} $ & $ AP^{m} $ & $ AP^{m}_{50} $  & $ AP^{m}_{75} $ \\
\midrule
ResNet18 ~\cite{resnet} & 31.2M & 36.9 & 57.1 & 40.0 & 33.6 & 53.9 & 35.7 \\
PVT-T ~\cite{wang2021pyramid} & 32.9M & 39.8 & 62.2 & 43.0 & 37.4 & 59.3 & 39.9 \\
PVTv2-B1 ~\cite{wang2021pvtv2} & 33.7M & 41.8 & 64.3 & 45.9 & 38.8 & 61.2 & 41.6 \\
\textbf{HLG-Tiny} & {30.6M} & {\textbf{44.7}} & {67.1} & {48.8} & {{\bf 40.9}} & {63.7} & {44.2} \\
\midrule
ResNet50 ~\cite{resnet} & 44.2M & 38.0 & 58.6 & 41.4 & 34.4 & 55.1 & 36.7 \\
PVT-S ~\cite{wang2021pyramid} & 44.1M & 40.4 & 62.9 & 43.8 & 37.8 & 60.1 & 40.3 \\
ViL-S ~\cite{zhang2021multi} & 45.0M & 41.8 & 64.1 & 45.1 & 38.5 & 61.1 & 41.4 \\
Swin-T ~\cite{liu2021swin} & 47.8M & 42.2 & 64.6 & 46.2 & 39.1 & 61.6 & 42.0 \\
RegionViT-S+ ~\cite{chen2021regionvit} & 50.9M & 44.2 & 67.3 & 48.2 & 40.8 & 64.1 & 44.0 \\
\textbf{HLG-Small} & {43.7M} & {\textbf{46.0}} & {68.6} & {50.7} & {\textbf{42.0}} & {65.7} & {45.1}  \\
\midrule
ResNet101 ~\cite{resnet} & 63.2M & 40.4 & 61.1 & 44.2 & 36.4 & 57.7 & 38.8 \\
ResNeXt101~\cite{xie2017aggregated} & 62.8M & 41.9  & 62.5 & 45.9  & 37.5  & 59.4 & 40.2 \\
PVT-M ~\cite{wang2021pyramid} & 63.9M & 42.0 & 64.4 & 45.6 & 39.0 & 61.6 & 42.1 \\
ViL-M ~\cite{zhang2021multi} & 60.1M & 43.4 & 65.9 & 47.0 & 39.7 & 62.8 & 42.1 \\
Swin-S ~\cite{liu2021swin} & 69.1M & 44.8 & 66.6 & 48.9 & 40.9 & 63.4 & 44.2 \\
\textbf{HLG-Medium} & {67.3M} & {\textbf{48.2}} & {70.2} & {53.0} & {{\bf 43.4}} & {67.0} & {46.7}  \\
\midrule
ResNeXt101 ~\cite{xie2017aggregated} & 101.9M & 42.8 & 63.8 & 47.3 & 38.4 & 60.6 & 41.3 \\
PVT-L ~\cite{wang2021pyramid} & 81.0M & 42.9 & 65.0 & 46.6 & 39.5 & 61.9 & 42.5 \\
ViL-B ~\cite{zhang2021multi} & 76.1M & 45.1 & 67.2 & 49.3 & 41.0 & 64.3 & 44.2 \\
Swin-B ~\cite{liu2021swin} & 107.2M & 45.5 & - & - & 41.3 & - & - \\
RegionViT-B+ ~\cite{chen2021regionvit} & 93.2M & 46.3 & 69.1 & 51.2 & 42.4 & 66.2 & 45.6 \\
\textbf{HLG-Large} & {103.6M} & {\textbf{49.1}} & {71.0} & {53.9} & {{\bf44.2}} & {68.0} & {47.8}  \\
\bottomrule
\end{tabular*}
\end{table*}

Table~\ref{tab:hlg_detection_maskrcnn} reports the bounding box AP and mask box AP of the Mask R-CNN detection framework \cite{he2017mask} in different backbones.
We make similar observations.
Our HLG Transformer outperforms all other alternatives consistently. 
Concretely, HLG-Tiny significantly surpasses ResNet-18 by 7.8\%, PVT-Tiny by 4.9\% and PVT-v2-B1 by 2.9\% in AP, using similar parameters. 
Moreover, HLG-Medium surpasses ResNet-101 by 7.8\%, PVT-M by 4.8\% and Swin-S by 3.4\% using similar parameters. 
Similar performance margins are achieved by
HLG-Small and HLG-Large over the competitors.

\noindent \textbf{Qualitative evaluation}

\begin{figure}[t]
  \centering
  \subfloat[Detection results of PVT~\cite{wang2021pyramid}]{
    \includegraphics[width=1.0\linewidth]{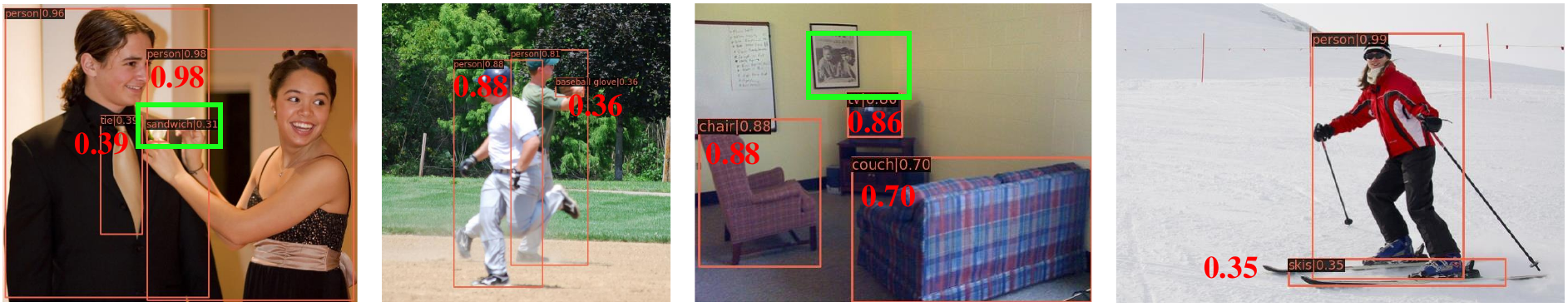}
    \label{detection_pvt}
  } \\ 
  \subfloat[Detection results of HLG]{
    \includegraphics[width=1.0\linewidth]{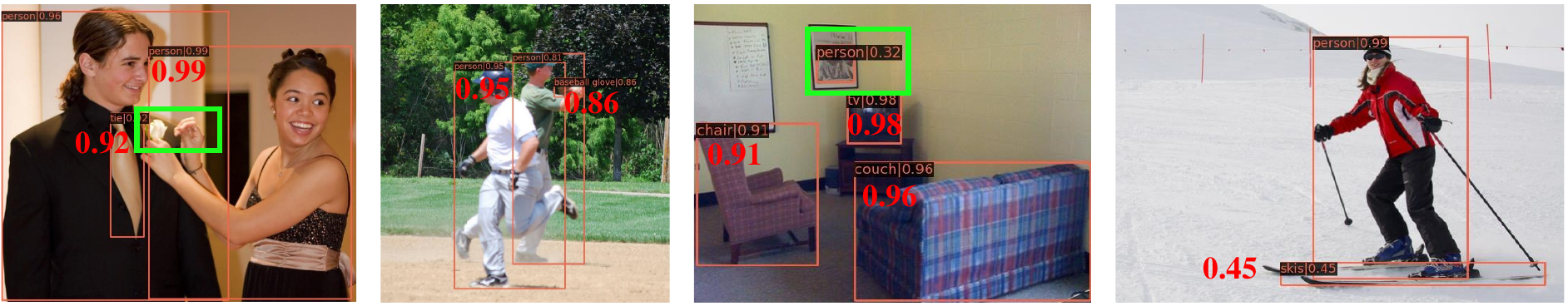}
    \label{detection_hlg}
  }
  \caption{Qualitative object detection results on COCO 2017. The RetinaNet \cite{lin2017focal} is used as the framework with (a) PVT-Medium \cite{wang2021pyramid} and (b) HLG-Medium as the backbone in comparison.}
  \label{fig:detection_result_vis}
\end{figure}


    


\begin{figure}[t]
  \centering
  \subfloat[Segmentation results of PVT~\cite{wang2021pyramid}]{
    \includegraphics[width=1.0\linewidth]{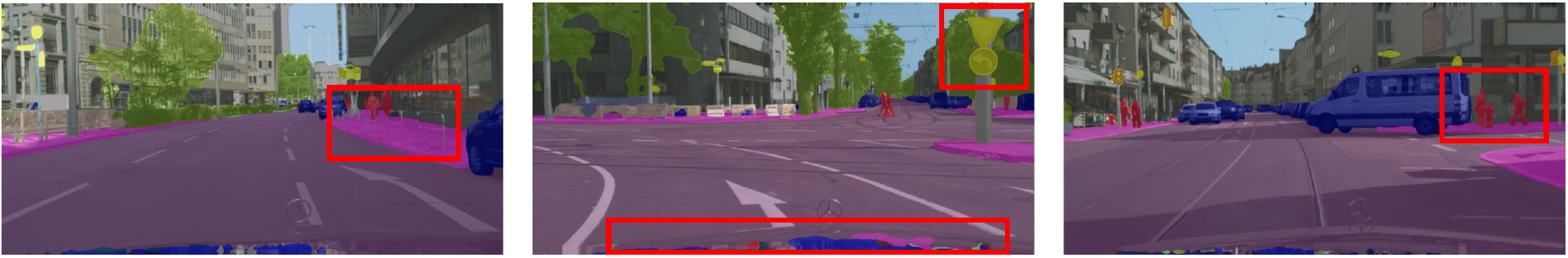}
    \label{segmentation_setr}
  } \\ 
  \subfloat[Segmentation results of HLG]{
    \includegraphics[width=1.0\linewidth]{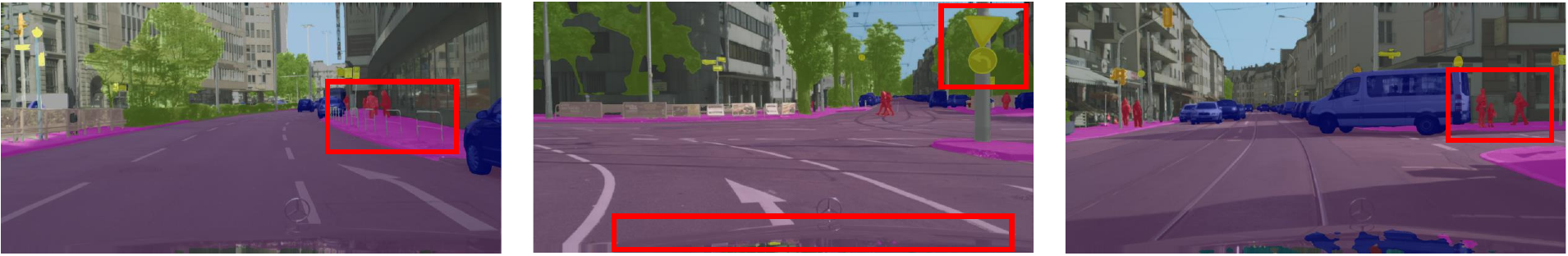}
    \label{segmentation_hlg}
  }
  \caption{Qualitative semantic segmentation results on Cityscapes.
  The UperNet is used as the framework with (a) PVT-Medium \cite{wang2021pyramid} and (b) HLG-Medium as the backbone in comparison.
  }
  \label{fig:segmentation_result_vis}
\end{figure}

We further provide qualitative evaluation by visualizing the results of object detection and semantic segmentation.
We compare our HLG-Medium and PVT-M with similar parameters.
For object detection, we use RetinaNet as the framework on COCO.
As shown in Figure \ref{fig:detection_result_vis},
it is clear that HLG Transformer is superior in detecting small objects,
thanks to its ability of better perceiving intra- and inter-patch
in visual representation learning.
For semantic segmentation, we use the UperNet as the framework
on Cityscapes.
It is shown in Figure \ref{fig:segmentation_result_vis}
that our HLG Transformer can better segment small object instances.
For example, in the first sample, HLG Transformer can infer the segmentation of pedestrians and railings more accurately. In the second sample, the traffic sign can be segmented better by HLG. In the last example, PVT cannot tell apart from the pedestrians and background around the boundary as well as HLG.



\begin{table}[!hbt]
\caption{\textbf{Ablation on window embedding: } Performance on ImageNet classification and COCO object detection under different window embedding approaches. ``W.E." means window embedding, ``DWConv" means depth-wise convolution~
\cite{chollet2017xception}}
\label{tab:ablation_we_coco}
\centering
\begin{tabular*}{\columnwidth}{@{}lcccc@{}}
\toprule
Method & W.E. & \#Params & Top-1~(\%) & AP\\
\midrule
HLG-Tiny & DWConv  & 11.0M & \textbf{81.14} & 39.6\\
HLG-Tiny & Avgpool  & 11.0M & 81.11 & \textbf{42.4}\\
\bottomrule
\end{tabular*}
\end{table}
\begin{table}[!htb]
\caption{\textbf{Ablation on dilated local distance attention mechanism: } Performance on ImageNet classification under different local attention setting. ``Plain+plain" denotes that only plain local attention are used in HLG transformer layers while ``plain+dila" denotes that plain and dilated local attention are hierarchically applied.
{The number after ``plain+dila'' denotes the dilation rate.}
``W.E.'' means window embedding, while ``DWConv'' means depth-wise convolution~
\cite{chollet2017xception}.}
\label{tab:ablation_hier}
\centering
\begin{tabular*}{\columnwidth}{@{}lcccc@{}}
\toprule
Method & W.E.  & Hierarchical & \#Params & Top-1~(\%)\\
\midrule
HLG-Tiny & Avgpool  & plain+plain & 11.0M & 80.58\\
{HLG-Tiny} & {Avgpool}  & {plain+dila 7} & {11.0M} & {81.11}\\
{HLG-Tiny} & {Avgpool}  & {plain+dila 5} & {11.0M} & {80.77}\\
{HLG-Tiny} & {Avgpool}  & {plain+dila 8} & {11.0M} & {81.08}\\
\bottomrule
\end{tabular*}
\end{table}
\begin{table}[!hbt]
\caption{{\textbf{Step by step ablation study.} We use PVT~\cite{wang2021pyramid} as the baseline but use our architecture specification in Table~\ref{tab:hlg_architecture}. Term ``shared local" denotes shared local attention and ``plain+dilated" denotes the plain and dilated local attnetion are hierarchically applied.
}
}
\label{tab:ablation_stepbystep}
\centering
\begin{tabular*}{\columnwidth}{@{}lcccc@{}}
\toprule
Method & \#Params & FLOPs & FPS & Top-1~(\%) \\
\midrule
{Baseline} & {24.5M} & {4.0G} & {802} & {79.2}\\
{+ DWMLP} & {24.2M} & {4.2G}  & {793} & {80.3}\\
{++ shared local} & {24.2M} & {4.7G} & {697} & {81.9}\\
{+++ plain+dilated} & {24.2M} & {4.7G} & {697} & {82.3}\\
\bottomrule
\end{tabular*}
\end{table}
\subsection{Ablation studies on HLG}

We evaluate two window embedding methods (depth-wise convolution~\cite{chollet2017xception} and average pooling).
For both, we set kernel size and stride as window size.
Table~\ref{tab:ablation_we_coco} shows that the two methods 
perform similarly on image classification whilst average pooling is better for object detection.
One reason is that average pooling can generalize over different 
kernel sizes flexibly.

Table~\ref{tab:ablation_hier} shows effect of hierarchical local attention mechanism. 
It is observed that the application of hierarchical local attention mechanism brings 0.64\% performance increment.

With PVT as the baseline, Table~\ref{tab:ablation_stepbystep} validates the effect of our key components
on ImageNet-1K.
Specifically, by adding depth-wise convolution to MLP of Transformer layer and substituting patch embedding with strided depth-wise convolution in MLP, we cut down the parameter size at a cost of slight FLOPs, bringing 1.1\% boost.
With local attention and sharing parameter/computation over local and global attention,
a gain of 1.0\% accuracy is obtained 
with the same parameter size and an admissible FLOPs increment.
Hierarchical local attention further yields a gain of 0.4\%.

\section{Conclusion}

{
In this work, we have presented an alternative 
perspective for visual representation learning of dense visual prediction tasks
by exploiting the ViT architectures.
Instead of gradually increasing the receptive field with CNN models,
we made a step change at the {\em architectural} level to elegantly solve the limited receptive field challenge.
We implemented the proposed idea with Transformers
capable of modeling global context at every stage of representation learning.
For tackling general dense visual prediction tasks in a cost-effective manner,
we further introduce Hierarchical Local Global (HLG) Transformers with local visual priors and global context learning integrated coherently in a unified architecture. 
Extensive experiments demonstrate that our models outperform existing alternatives on several visual recognition tasks.
}

\section{Data Availability Statement}
The datasets generated during and/or analysed during the current study are available in the Imagenet~\cite{imagenet} (\url{https://www.image-net.org/}), ImageNet-v2~\cite{recht2019imagenet} (\url{https://github.com/modestyachts/ImageNetV2}), COCO~\cite{COCO_dataset} (\url{https://cocodataset.org}), ADE20K~\cite{ADE20K} (\url{https://groups.csail.mit.edu/vision/datasets/ADE20K/}), Cityscapes~\cite{cordts2016cityscapes} (\url{https://www.cityscapes-dataset.com}), Pascal Context~\cite{mottaghi_cvpr14} (\url{https://cs.stanford.edu/~roozbeh/pascal-context/}) repositories.

\section*{Acknowledgments}
We thank Hengshuang Zhao, Zekun Luo and Yabiao Wang for valuable discussions.
This work was supported in part by STI2030-Major Projects (Grant No. 2021ZD0200204), National Natural Science Foundation of China (Grant No. 62106050 and 62376060),
Natural Science Foundation of Shanghai (Grant No. 22ZR1407500), 
USyd-Fudan BISA Flagship Research Program and Lingang Laboratory (Grant No. LG-QS-202202-07).
\begin{appendix}
\section{Visualizations}

\subsection{Position Embedding}
\begin{figure}[h!]\centering
\includegraphics[width=0.92\linewidth]{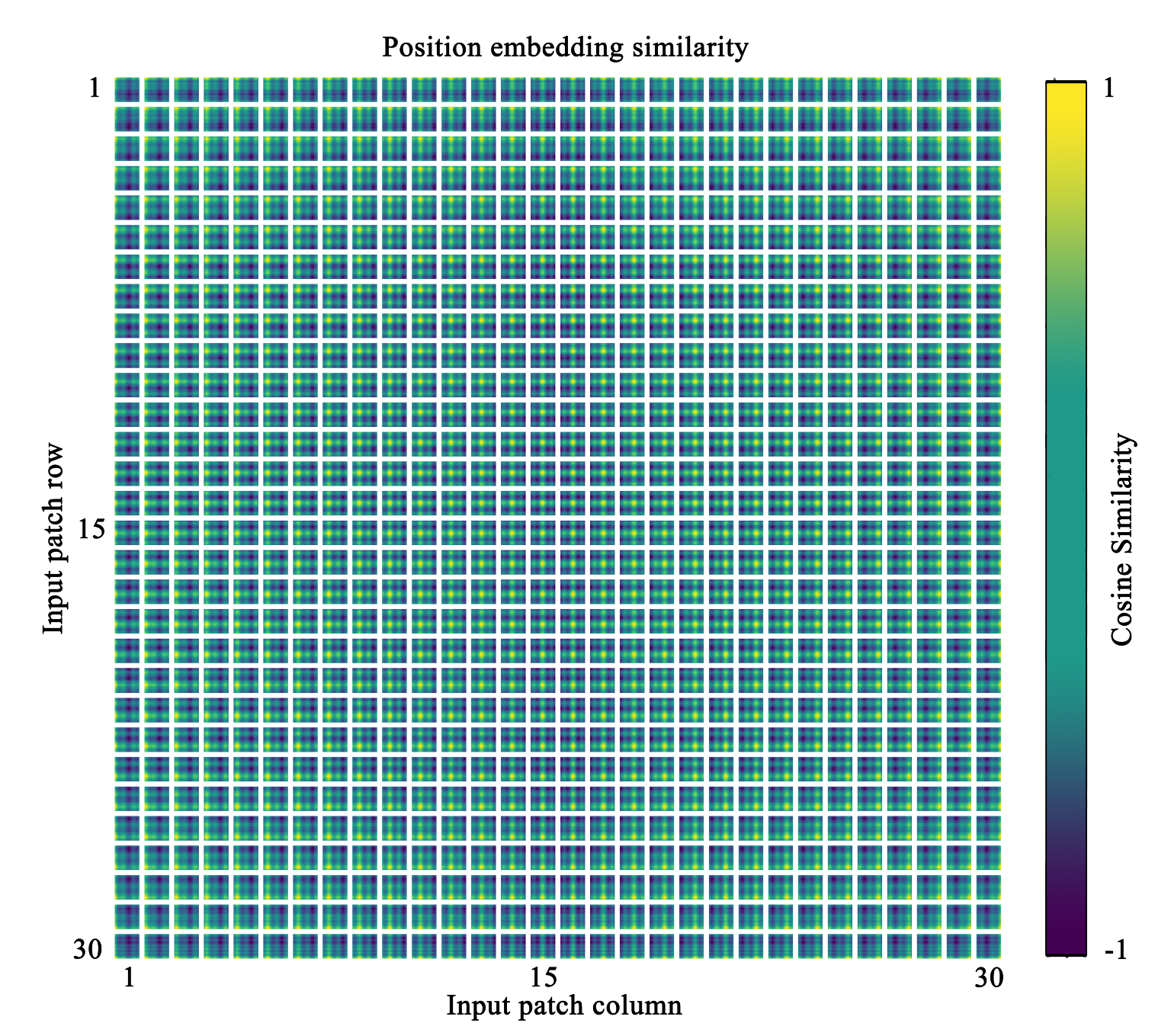}
\caption{Similarity of position embeddings of~\pupModel~trained on Pascal Context.
Tiles show the cosine similarity between the position embedding of the patch with the indicated row and column and the position embeddings of all other patches.}
\label{fig:pos}
\end{figure}

Visualization of the learned position embedding in Figure~\ref{fig:pos} shows that the model learns to encode distance within the image in the similarity of position embeddings.

\subsection{Features}
 
\begin{figure*}[!ht]\centering
\includegraphics[width=0.85\linewidth]{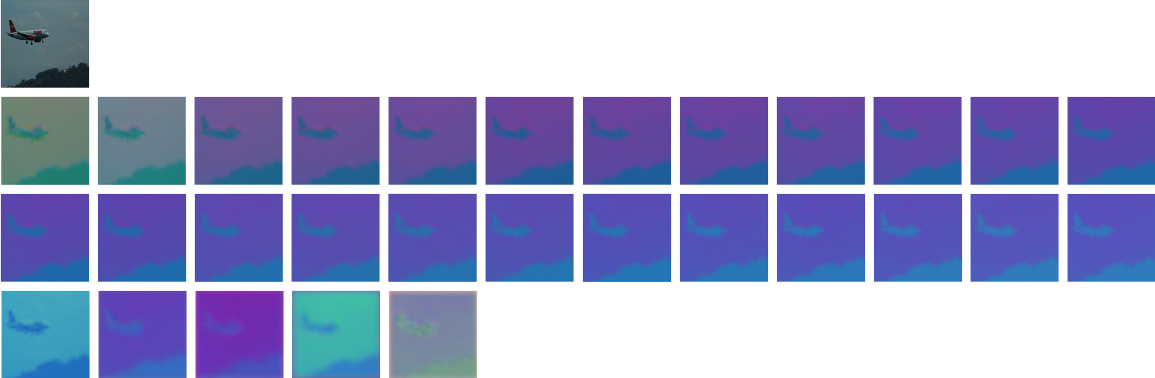}
\caption{Visualization of output feature of layer $Z^1-Z^{24}$ and $U^1-U^5$ of \pupModel~trained on Pascal Context.
Best view in color.
\textbf{First row:} The input image. 
\textbf{Second row:} Layer $Z^1$-$Z^{12}$.
\textbf{Third row:} Layer $Z^{13}$-$Z^{24}$.
\textbf{Fourth row:} Layer $U^1-U^5$.
} 
\label{fig:layer-pup}
\end{figure*}

Figure~\ref{fig:layer-pup} shows the feature visualization of our \pupModel.
For the encoder, 24 output features from the 24 Transformer layers namely $Z^1-Z^{24}$ are collected. Meanwhile, 5 features ($U^1-U^5$) right after each bilinear interpolation in the decoder head are visited.

\subsection{Attention Maps}

\begin{figure}[ht]\centering
\includegraphics[width=0.75\linewidth]{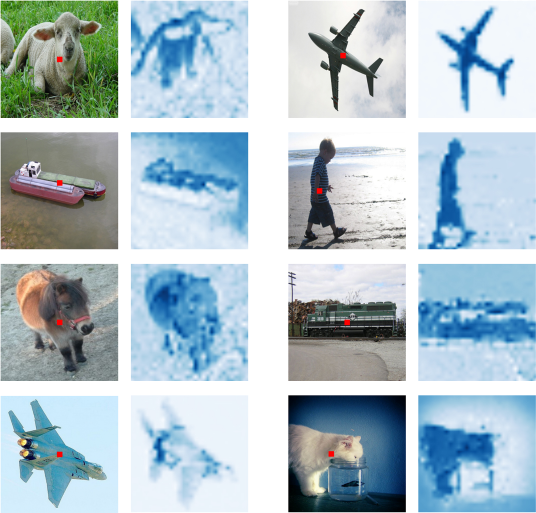}
\caption{The first and third columns show images from Pascal Context.
The second and fourth columns illustrate the attention map of the picked points (red).}
\label{fig:ori-gt-att}
\end{figure}

\begin{figure}[t]\centering
\includegraphics[width=1.02\linewidth]{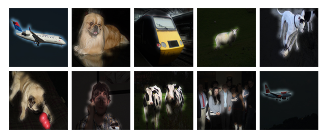}
\vspace{-0.3cm}
\caption{
Examples of attention maps from SETR trained on Pascal Context. 
}
\label{fig:attention}
\end{figure}

\begin{figure*}[ht!]\centering
\includegraphics[width=0.85\linewidth]{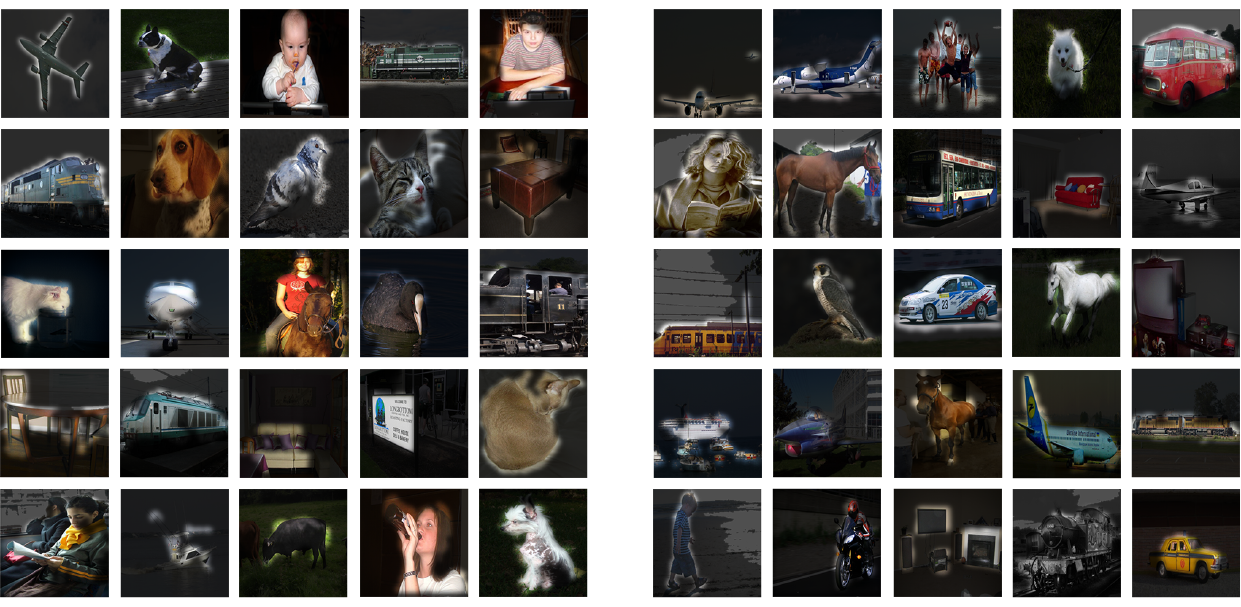}
\caption{More examples of attention maps from~\pupModel~trained on Pascal Context. }
\label{fig:attention-50}
\end{figure*}

Attention maps (Figure~\ref{fig:attention} and Figure~\ref{fig:attention-50}) in each Transformer layer catch our interest. 
There are 16 heads and 24 layers in T-large.
Similar to \cite{abnar2020quantifying}, a recursion perspective into this problem is applied. 
Figure~\ref{fig:ori-gt-att} shows the attention maps of different selected spatial points (red).


\end{appendix}


\clearpage
\bibliographystyle{spbasic}      
\bibliography{main}   


\end{document}